\documentclass[pdflatex,sn-mathphys-num]{sn-jnl}

\usepackage{graphicx}%
\usepackage{multirow}%
\usepackage{amsmath,amssymb,amsfonts}%
\usepackage{amsthm}%
\usepackage{mathrsfs}%
\usepackage[title]{appendix}%
\usepackage{xcolor}%
\usepackage{textcomp}%
\usepackage{manyfoot}%
\usepackage{booktabs}%
\usepackage{algorithm}%
\usepackage{algorithmicx}%
\usepackage{algpseudocode}%
\usepackage{listings}%
\usepackage{array}
\usepackage{colortbl}
\usepackage{booktabs}
\usepackage{tabularx}
\PassOptionsToPackage{hyphens}{url}\usepackage{hyperref}
\usepackage{afterpage} 
\usepackage{subcaption}
\usepackage{geometry}
\usepackage{mwe} 
\usepackage[english]{babel}
\usepackage[T1]{fontenc}
\usepackage[
    left = \flqq{},%
    right = \frqq{},%
    leftsub = \flq{},%
    rightsub = \frq{} %
]{dirtytalk}
\let\cline\cmidrule

\theoremstyle{thmstyleone}%
\theoremstyle{thmstyletwo}%

\theoremstyle{thmstylethree}%

\begin{document}

\title[Comparative Study of Zero-Shot Cross-Lingual Transfer for Bodo POS and NER Tagging Using Gemini 2.0 Flash Thinking Experimental Model]{Comparative Study of Zero-Shot Cross-Lingual Transfer for Bodo POS and NER Tagging Using Gemini 2.0 Flash Thinking Experimental Model}


\author*[1,2]{\fnm{Sanjib} \sur{Narzary}}\email{narzary@iitg.ac.in}\email{san@cit.ac.in}

\author[2]{\fnm{Bihung} \sur{Brahma}}\email{b.brahma@cit.ac.in}
\equalcont{These authors contributed equally to this work.}

\author[2]{\fnm{Haradip Kumar} \sur{Mahilary}}\email{hk.mahilary@cit.ac.in}
\equalcont{These authors contributed equally to this work.}
\author[2]{\fnm{Mahananda} \sur{Brahma}}\email{mn.brahma@cit.ac.in}
\equalcont{These authors contributed equally to this work.}
\author[1]{\fnm{Bidisha} \sur{Som}}\email{bidisha@iitg.ac.in}
\author[1]{\fnm{Sukumar} \sur{Nandi}}\email{sukumar@iitg.ac.in}
\affil*[1]{\orgdiv{Centre for Linguistic Science and Technology}, \orgname{IIT Guwahati}, \orgaddress{\street{North Guwahati}, \city{Guwahati}, \postcode{781039}, \state{Assam}, \country{India}}}
\affil[2]{\orgname{Central Institute of Technology Kokrajhar}, \orgaddress{\street{JD Road}, \city{Kokrajhar}, \postcode{783370}, \state{Assam}, \country{India}}}


\abstract{Named Entity Recognition (NER) and Part-of-Speech (POS) tagging are critical tasks for Natural Language Processing (NLP), yet their availability for low-resource languages (LRLs) like Bodo remains limited. This article presents a comparative empirical study investigating the effectiveness of Google's Gemini 2.0 Flash Thinking Experiment model for zero-shot cross-lingual transfer of POS and NER tagging to Bodo. We explore two distinct methodologies: (1) direct translation of English sentences to Bodo followed by tag transfer, and (2) prompt-based tag transfer on parallel English-Bodo sentence pairs. Both methods leverage the machine translation and cross-lingual understanding capabilities of Gemini 2.0 Flash Thinking Experiment to project English POS and NER annotations onto Bodo text in CONLL-2003 format. Our findings reveal the capabilities and limitations of each approach, demonstrating that while both methods show promise for bootstrapping Bodo NLP, prompt-based transfer exhibits superior performance, particularly for NER.  We provide a detailed analysis of the results, highlighting the impact of translation quality, grammatical divergences, and the inherent challenges of zero-shot cross-lingual transfer.  The article concludes by discussing future research directions, emphasizing the need for hybrid approaches, few-shot fine-tuning, and the development of dedicated Bodo NLP resources to achieve high-accuracy POS and NER tagging for this low-resource language.}




\maketitle

\section{Introduction and Motivation}

\subsection{The Significance of POS and NER Tagging in NLP}
Part-of-Speech (POS) tagging and Named Entity Recognition (NER) are fundamental tasks within the field of Natural Language Processing (NLP), serving as essential prerequisites for a multitude of downstream applications. POS tagging, the process of assigning grammatical categories to individual words within a sentence (e.g., noun, verb, adjective, adverb), provides crucial syntactic information that underpins higher-level language understanding. NER, on the contrary, focuses on identifying and classifying named entities – real-world objects that are designated with a proper name – into predefined semantic categories such as persons, organizations, locations, dates, times, and quantities \citep{jbp:/content/journals/10.1075/li.30.1.03nad, 10.1145/3522593}.

The synergy of POS and NER tagging empowers a wide spectrum of NLP applications. In information extraction, NER helps to pinpoint key entities, while POS tags help to understand the relationships between these entities and other words in the text, facilitating the extraction of structured information from unstructured text \citep{lample-etal-2016-neural}. Machine translation systems benefit from POS tagging to improve syntactic analysis and word order prediction, and NER to ensure accurate translation of named entities in languages \citep{NIPS2017_3f5ee243_transformer_vaswani}. Question-answer systems rely on both NER and POS to understand the question's intent, identify relevant entities and relationships in the knowledge base, and formulate accurate answers. Text summarization algorithms leverage NER to identify salient entities and POS tags to preserve grammatical coherence and readability in summaries. The precision and robustness of these core NLP tasks directly impact the effectiveness of these downstream applications.

\subsection{NLP Challenges for Low-Resource Languages: The Case of Bodo}
Despite the remarkable advances in NLP, there is a significant disparity between high-resource languages (HRLs) and low-resource languages (LRLs). Although HRLs, such as English, Spanish, French, and Mandarin, benefit from a wealth of annotated data, sophisticated NLP tools, and substantial research investment, LRLs lag considerably behind \citep{joshi-etal-2020-state, mager2018challenges}. LRLs are characterized by a scarcity of digital resources, including annotated corpora, lexicons, grammars, and readily available NLP tools. This resource scarcity poses a significant bottleneck for developing NLP technologies for these languages, hindering their integration into the increasingly digital world.

Bodo, a vibrant Tibeto-Burman language spoken by approximately 1.5 million people mainly in the Bodoland Territorial Region of Northeast India, serves as a representative example of an LRL facing NLP resource limitations \citep{Narzary2024}. Although Bodo has a rich linguistic tradition and growing online presence, it lacks the extensive NLP infrastructure available for more widely spoken languages. Although preliminary efforts have explored deep learning approaches for part-of-speech tagging \citep{Pathak_Narzary_Nandi_Som_2025} and named entity recognition \citep{NARZARY20242405}, their reliance on manually annotated datasets presents a potential bottleneck. Specifically, robust and freely accessible POS and NER taggers for Bodo are not readily available.  Developing such taggers from scratch using traditional supervised learning methods would require the creation of large manually annotated datasets, a resource-intensive and time-consuming undertaking, particularly for languages with limited linguistic documentation and expert annotators \citep{palmer2010computational, bhat-varma-2023-large}. The absence of effective Bodo NER and POS taggers impedes progress in more advanced NLP applications for the Bodo language community, limiting access to the benefits of NLP technology.

\subsection{Harnessing Gemini 2.0 Flash Thinking Experiment model for Zero-Shot Cross-Lingual Transfer}
The emergence of Large Language Models (LLMs) \footnote{Large Language Model \url{https://paperswithcode.com/task/large-language-model}} has ushered in a paradigm shift in NLP, offering new possibilities to address the challenges of low-resource scenarios \citep{zhang2024affective}. LLMs, pre-trained on massive datasets encompassing multiple languages, exhibit remarkable cross-lingual transfer capabilities, enabling them to generalize knowledge learned from resource-rich languages to resource-scarce ones \citep{conneau-etal-2020-emerging, pires-etal-2019-multilingual}.  Google's Gemini 2.0 Flash Thinking Experimental model \footnote{Google Gemini 2.0 Flash Thinking \url{https://deepmind.google/technologies/gemini/flash-thinking/}} represents a leading example of such advanced LLMs, designed for efficient inference and adaptable performance across diverse tasks and languages.  It is hypothesized that the Gemini 2.0 Flash Thinking Experiment possesses a deep, cross-lingual understanding of language structures and semantics, making it a promising candidate for zero-shot cross-lingual transfer.

This article presents an empirical study evaluating the application of Google's Gemini 2.0 Flash Thinking Experiment model for zero-shot cross-lingual transfer of POS and NER tagging to Bodo. We explore two distinct methodologies to leverage Gemini 2.0 Flash Thinking Experiment's model capabilities:

1. \textbf{Method 1: Translation-Based Tag Transfer:} This approach utilizes Gemini 2.0 Flash Thinking Experiment's machine translation module to translate English sentences into Bodo. Subsequently, we investigate the direct transfer of POS and NER tags from the English source sentence to its Bodo translation. This method assesses the feasibility of using MT-induced word alignments and cross-lingual representations for tag projection.

2. \textbf{Method 2: Prompt-Based Tag Transfer:} This methodology directly prompts Gemini 2.0 Flash Thinking Experiment to perform both machine translation and tag transfer simultaneously. By providing parallel English-Bodo sentence pairs as input, we aim to guide the model not only translate but also to explicitly project and generate POS and NER tags for the Bodo sentence in a zero-shot manner. This approach explores the potential of LLMs for more direct and context-aware cross-lingual tag transfer.

By comparatively evaluating these two methods, we seek to gain insights into the effectiveness of Gemini 2.0 Flash Thinking Experiment for zero-shot cross-lingual POS and NER tagging for Bodo. The study aims to assess the model's capabilities, identify limitations, and provide a foundation for future research into more robust and resource-efficient NLP techniques for Bodo and other low-resource languages. The ultimate goal is to explore pathways towards democratizing NLP technology and making it accessible to language communities currently underserved by these advancements \citep{Narzary2024}.

\section{Literature Review}
\subsection{Evolution of POS and NER Tagging Techniques}
The tasks of Part-of-Speech (POS) tagging and Named Entity Recognition (NER) have been central to Natural Language Processing (NLP) research for decades, witnessing a significant evolution in methodologies and performance. Early approaches to both tasks relied heavily on rule-based systems and manually crafted features, often requiring substantial linguistic expertise and language-specific resources \citep{mccallum2000maximum, ratnaparkhi1996maximum, brill-1995-transformation}. Rule-based POS taggers, for example, employed handwritten rules to disambiguate word categories based on contextual patterns and morphological cues. Similarly, early NER systems often utilized gazetteers, handcrafted dictionaries of named entities, and rule-based patterns to identify and classify entities in text \citep{jbp:/content/journals/10.1075/li.30.1.03nad}. While these rule-based systems offered interpretability and precision in specific domains, they suffered from limitations in scalability, robustness, and adaptability to new domains and languages.

The rise of statistical NLP marked a significant shift towards data-driven approaches. Hidden Markov Models (HMMs) \citep{eddy1996hidden} and Conditional Random Fields (CRFs) \citep{sutton2012introduction} emerged as prominent statistical frameworks for both POS tagging and NER \citep{toutanova2003feature, mccallum2000maximum, ratnaparkhi1996maximum}. HMM-based POS taggers, for example, modeled POS tagging as a sequence labeling problem, leveraging probabilistic models to predict the most likely POS tag sequence for a given sentence. CRFs, a discriminative graphical model, offered advantages over HMMs by allowing more flexible feature engineering and capturing long-range dependencies in the input text \citep{lafferty2001conditional}. Statistical methods significantly improved the accuracy and robustness of POS and NER systems compared to rule-based approaches, but still relied on manually engineered features and task-specific knowledge.

The deep learning revolution in NLP has led to a paradigm shift in POS and NER tagging, achieving state-of-the-art performance, and reducing the reliance on manual feature engineering. Recurrent neural networks (RNNs) \citep{bi-rnn}, particularly long-short-term memory networks (LSTMs) \citep{hochreiter1997long} and Gated Recurrent Units (GRUs) \citep{cho2014learningphraserepresentationsusing, chung2014empiricalevaluationgatedrecurrent}, proved highly effective in capturing sequential dependencies in text, enabling models to learn contextual representations of words for sequence labeling tasks \citep{huang2015bidirectionallstmcrfmodelssequence, ma-hovy-2016-end, reimers-gurevych-2017-reporting, akbik-etal-2018-contextual}. Bi-directional LSTMs \citep{huang2015bidirectionallstmcrfmodelssequence}, which process input sequences in both forward and backward directions, further enhanced performance by capturing contextual information from both past and future words. The integration of Convolutional Neural Networks (CNNs) \citep{oshea2015introductionconvolutionalneuralnetworks} into hybrid architectures, such as BiLSTM-CNN-CRF \citep{ma2016endtoendsequencelabelingbidirectional} models, allowed for the simultaneous learning of character-level and word-level representations, improving the handling of morphological variations and out-of-vocabulary words \citep{ma-hovy-2016-end}.

The Transformer architecture, introduced by \citep{NIPS2017_3f5ee243_transformer_vaswani}, has further revolutionized NLP, establishing a new state of the art for a wide range of tasks, including POS and NER tagging. Transformer-based models, such as BERT (Bidirectional Encoder Representations from Transformers) \citep{devlin-etal-2019-bert} and its multilingual variants (mBERT \citep{pires-etal-2019-multilingual}, XLM-RoBERTa \citep{conneau-etal-2020-unsupervised}), leverage self-attention mechanisms to capture long-range dependencies and contextual information more effectively than RNNs, while also enabling highly parallelized training. Pre-trained on massive text corpora, Transformer models learn rich contextualized word embeddings that can be fine-tuned for specific downstream tasks, achieving remarkable performance gains, especially in data-scarce scenarios and cross-lingual settings \citep{devlin-etal-2019-bert, conneau2019cross, pires-etal-2019-multilingual}.

\subsection{Cross-Lingual Transfer Learning and the Role of Machine Translation}
Cross-lingual transfer learning has emerged as a crucial paradigm in NLP, aiming to bridge the performance gap between high-resource and low-resource languages. The core idea behind transfer learning is to leverage knowledge acquired from a source language (typically resource-rich) to improve model performance on a target language (typically resource-scarce) \citep{pan2010survey, ruder2017overviewmultitasklearningdeep, ZHAO2024122807}. In the cross-lingual context, this involves transferring linguistic knowledge across languages, capitalizing on shared linguistic structures and universal language representations.

Several techniques have been developed for cross-lingual transfer learning in NLP \citep{conneau2019cross, lample2019crosslinguallanguagemodelpretraining}. Cross-lingual word embeddings represent one prominent approach, focusing on mapping word embeddings from different languages into a shared vector space \citep{mikolov2013distributed, mikolov2013exploitingsimilaritieslanguagesmachine, faruqui-dyer-2014-improving}. By aligning word embeddings across languages, these techniques enable the transfer of semantic and syntactic knowledge, allowing models trained on one language to generalize to another. Multilingual models, such as mBERT \citep{pires-etal-2019-multilingual} and XLM-RoBERTa \citep{conneau-etal-2020-unsupervised}, take a more holistic approach by training a single model on data from multiple languages simultaneously \citep{devlin-etal-2019-bert, conneau2019cross}. These models learn language-invariant representations that capture universal linguistic features, facilitating zero-shot cross-lingual transfer, where a model trained on source languages can be directly applied to target languages without explicit fine-tuning on target language data.

Machine Translation (MT) \citep{WANG2022143} plays a multifaceted role in cross-lingual transfer learning.  Firstly, MT can be used as a data augmentation technique, translating labeled data from a high-resource source language to a low-resource target language to create pseudo-labeled training data \citep{yarowsky-1995-unsupervised, ruder2017overviewmultitasklearningdeep}. This approach, known as "translate-train," allows for leveraging existing annotations in resource-rich languages to bootstrap model training for LRLs. Secondly, MT systems, particularly advanced Neural Machine Translation (NMT) \citep{bahdanau2014neural, cho2014properties} models, can be integrated into cross-lingual transfer architectures. Attention mechanisms in NMT models, for instance, can provide word alignment information that can be exploited for tag projection, transferring annotations from source to target language sentences based on word-level correspondences \citep{bahdanau2014neural}.  Moreover, the cross-lingual representations learned by multilingual NMT models can be directly utilized for downstream NLP tasks in a zero-shot or few-shot transfer setting \citep{lample2019crosslinguallanguagemodelpretraining}.

Rapid progress in neural machine translation as significantly improved the feasibility and effectiveness of MT-based cross-lingual transfer. NMT models, especially Transformer-based architectures, have achieved near-human-level translation quality for many language pairs, providing high-fidelity translations that preserve semantic and syntactic information \citep{NIPS2017_3f5ee243_transformer_vaswani}. This advancement makes MT a powerful tool for bridging the resource gap between HRLs and LRLs, enabling the development of NLP technologies for languages with limited annotated data.

\subsection{Addressing the NLP Resource Gap for Bodo}
The Bodo language, representative of many LRLs worldwide, faces a significant NLP resource gap. The lack of annotated datasets, robust NLP tools, and dedicated research efforts hinders the development of effective NLP applications for the Bodo language community. While some initial steps have been taken in Bodo NLP, such as the development of morphological analyzers \citep{smit-etal-2014-morfessor, morphological-analyzer-22}  and machine translation systems \citep{narzary_attention_19, Narzary2024}, the crucial tasks of POS and NER tagging remain largely unaddressed.

Traditional supervised learning approaches for POS and NER tagging are data-hungry \citep{bhattacharjee2019named}, requiring substantial amounts of manually annotated text – a resource that is not readily available for Bodo.  This limitation necessitates the exploration of alternative, resource-efficient techniques, such as cross-lingual transfer learning. Zero-shot cross-lingual transfer, leveraging the cross-lingual capabilities of advanced LLMs like Gemini 2.0 Flash Thinking Experimental model, offers a promising pathway to bootstrap Bodo NLP without requiring extensive Bodo-specific annotations. By effectively transferring knowledge from resource-rich languages like English, we can potentially create initial POS and NER taggers for Bodo, paving the way for further NLP development and language technology access for the Bodo language community. This article aims to empirically investigate the potential of this zero-shot transfer approach, contributing to the broader goal of democratizing NLP and extending its benefits to all languages, regardless of their resource availability.

\section{Methodology}
\subsection{Architecture of Gemini 2.0 Flash Thinking Experiment for Cross-Lingual Tagging}
This study empirically evaluates the zero-shot cross-lingual transfer performance of Google's Gemini 2.0 Flash Thinking Experimental model \citep{gemini2} for Bodo POS and NER tagging, focusing on two distinct methodologies designed to leverage the model's capabilities in machine translation and cross-lingual representation.

\begin{figure}
    \centering
    \includegraphics[width=0.98\linewidth]{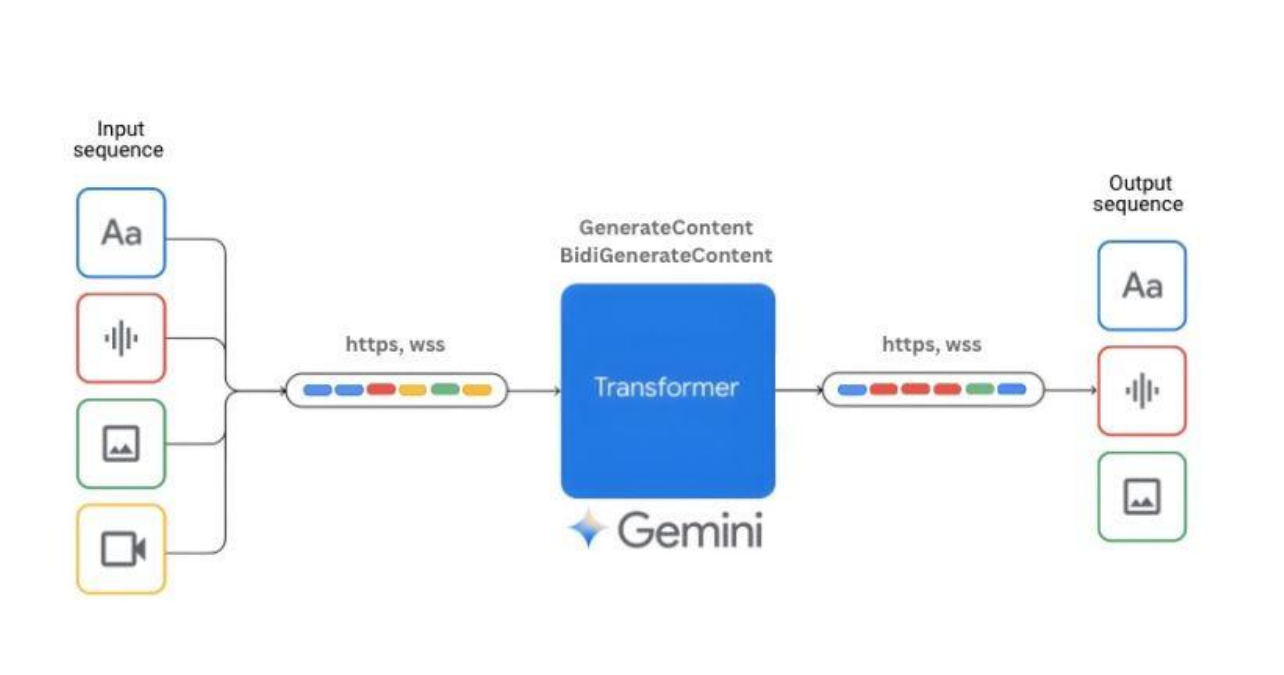}
    \caption[Gemini Architecture]{Abstract view of Google Gemini Architecture. Image Credit \footnotemark}
    \label{fig:gemini-architecture}
\end{figure}

Gemini 2.0 Flash Thinking Experiment, a state-of-the-art LLM, is hypothesized to possess a sophisticated architecture enabling effective zero-shot cross-lingual transfer \citep{gemini2}.  While the precise internal architecture is proprietary, for the purpose of this study, we assume a Transformer-based architecture, similar to models like mBERT and XLM-RoBERTa, pre-trained on a massive multilingual corpus \citep{devlin-etal-2019-bert, conneau2019cross}. This pre-training is assumed to have instilled in the model a rich, language-agnostic understanding of linguistic structures, semantic relationships, and cross-lingual word alignments.  We further assume that Gemini 2.0 Flash Thinking Experimental model \citep{gemini2} incorporates a dedicated module or mechanism for machine translation, enabling high-quality translations between a wide range of languages, including English and Bodo.

\footnotetext{Gemini Architecture  \url{https://tinyurl.com/4uhuuuu3}}

Based on these assumptions, we designed two distinct methodologies to evaluate Gemini 2.0 Flash Thinking Experiment's zero-shot cross-lingual tag transfer capabilities for Bodo POS and NER tagging:

\noindent
\textbf{Method 1: Translation-Based Tag Transfer}

1.  \textbf{English Sentence Input:} The input to the system is an English sentence from a general-domain corpus.

2.  \textbf{English POS and NER Tagging (Source Annotation):} The English sentence is pre-processed and annotated with POS and NER tags using spaCy's `en\_core\_web\_sm` model \citep{honnibal2020spacy}. These English tags serve as the source annotations for cross-lingual transfer.

3.  \textbf{Machine Translation (Gemini 2.0 Flash Thinking Experiment):} The machine translation module of the Gemini 2.0 Flash Thinking Experimental model is used to translate English sentencess into Bodo. The model's zero-shot translation capabilities are leveraged to generate Bodo translations without explicit fine-tuning on English-Bodo parallel data.

\begin{figure}
    \centering
    \includegraphics[width=0.98\linewidth]{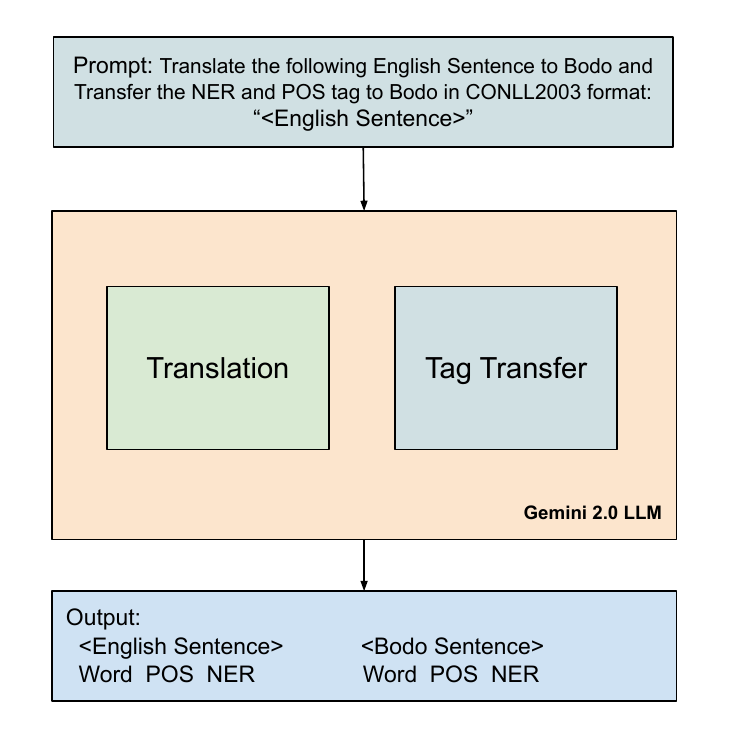}
    \caption{Prompt with translation based tag transfer architecture.}
    \label{fig:prompt-translate}
\end{figure}
4.  \textbf{Zero-Shot Tag Projection and Transfer:} The core of this method lies in the zero-shot projection and transfer of POS and NER tags from the English source sentence to the generated Bodo translation. Gemini 2.0 Flash Thinking Experiment, leveraging its cross-lingual understanding, implicitly performs word alignment during translation. Based on these implicit alignments, the system projects POS and NER tags from English words onto their corresponding Bodo words.  For words or phrases where direct alignment is ambiguous or absent, heuristic-based rules are applied to infer and transfer tags based on contextual similarity and grammatical function.
5.  \textbf{Bodo Tagged Sentence Output:} The final output is the Bodo sentence with projected POS and NER tags, formatted in CONLL-2003 column format.

\noindent
\textbf{Method 2: Prompt-Based Tag Transfer}

1.  \textbf{Parallel Sentence Pair Input:} This method takes pairs of parallel sentences as input, consisting of an English sentence and its corresponding Bodo translation (obtained from TDIL-DC Health and Tourism Domain).

2.  \textbf{Prompt Engineering for Gemini 2.0 Flash Thinking Experiment:} We employ a carefully designed prompt to instruct Gemini 2.0 Flash Thinking Experiment model to perform both POS and NER tagging for the Bodo sentence in a zero-shot fashion. The prompt provides the English sentence, its Bodo translation, and explicitly requests the model to output the Bodo sentence with POS and NER tags in CONLL-2003 format. The prompt leverages the model's instruction-following capabilities to guide it towards the desired task.  
\begin{figure}
    \centering
    \includegraphics[width=0.98\linewidth]{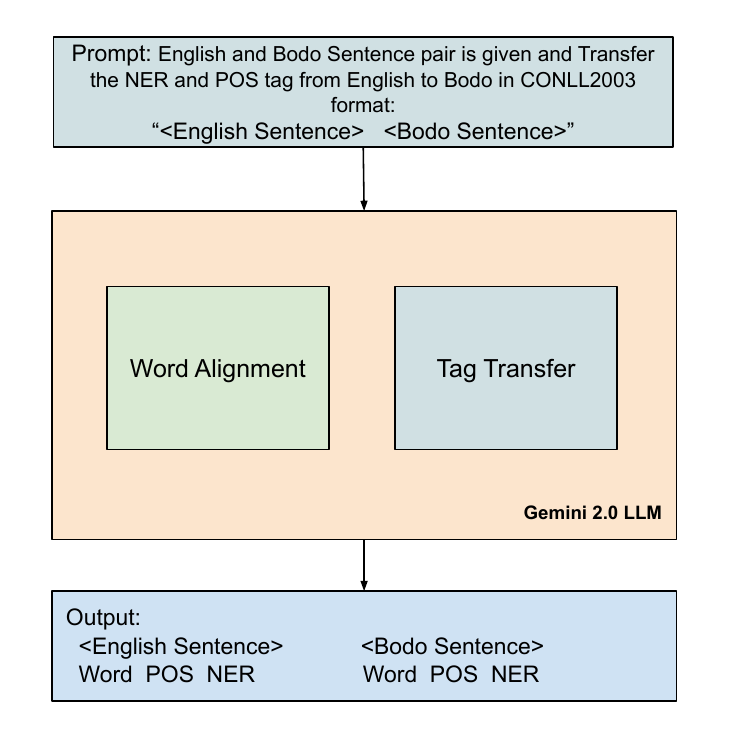}
    \caption{Parallel sentence based tag transfer architecture.}
    \label{fig:prompt-translate}
\end{figure}

\noindent
\textbf{Our Prompt Structure:} 
\say{Translate the following English sentence to Bodo and provide POS and NER tags for each Bodo word in CONLL-2003 format:

\say{English: [English Sentence]}

\say{Bodo Translation: [Bodo Sentence]}

\say{Tagged English:}

\say{Tagged Bodo:}
}

3.  \textbf{Zero-Shot Tag Generation (Gemini 2.0 Flash Thinking Experiment):} Gemini 2.0 Flash Thinking Experiment, guided by the prompt, processes the English-Bodo sentence pair and directly generates the tagged Bodo sentence. The model leverages its cross-lingual understanding and instruction-following abilities to perform both translation (implicitly to understand the Bodo sentence context) and zero-shot tag assignment.

4.  \textbf{Bodo Tagged Sentence Output:} The output is the Bodo sentence with POS and NER tags generated by Gemini 2.0 Flash Thinking Experiment model, formatted in CONLL-2003 column format.

\noindent
\textbf{Tag Transfer Mechanism within Gemini 2.0 Flash Thinking Experiment (Assumptions):}
For both methods, the underlying tag transfer mechanism within Gemini 2.0 Flash Thinking Experiment model is assumed to be a complex process leveraging the model's cross-lingual representations and attention mechanisms.  We hypothesize that the model performs implicit word alignment between English and Bodo sentences during translation, possibly through shared embedding spaces or cross-attention layers.  Tag projection is then guided by these alignments, prioritizing direct word-to-word mappings where available.  For cases where direct alignments are ambiguous or absent, the model likely leverages contextual information and learned cross-lingual syntactic patterns to infer and transfer tags, particularly for POS tagging.  For NER, the model may rely on its knowledge of named entity types in high-resource languages and attempt to generalize these categories to Bodo based on semantic similarity and contextual cues.  However, the precise details of this internal tag transfer mechanism remain a "black box" within the context of this experiment, as the Gemini 2.0 Flash Thinking Experiment model is a proprietary implementation of Google.

\subsection{Experimental Setup and Evaluation Metrics}

\noindent
\textbf{Dataset and Preprocessing:}
To ensure comparability, both Method 1 and Method 2 were tested on a shared dataset subset, which consisted of five randomly selected sentences from each of the health and tourism domains.  The English sentences are pre-processed using standard NLP techniques, including tokenization and sentence splitting, as performed by spaCy.  The corresponding Bodo sentences from the randomly selected test set were tokenized using the Bodo Tokenizer \footnote{Bodo Tokenizer \url{https://github.com/bodonlp/bodo-tokenizer}}. No additional preprocessing is applied to the Bodo text to maintain the focus on zero-shot transfer.

\noindent
\textbf{Evaluation Methodology:}
Due to the lack of a publicly available gold standard Bodo NER/POS dataset, we employ a qualitative evaluation methodology, focusing on manual assessment by a linguistic expert.  From the output of both Method 1 and Method 2, we randomly sample 5 tagged Bodo sentences each (10 sentences total).  A linguist fluent in Bodo and experienced in NLP is tasked with manually evaluating these sentences, assessing the accuracy and quality of the transferred POS and NER tags.

The qualitative evaluation focuses on the following criteria:

\begin{itemize}
    \item \textbf{POS Tag Accuracy:} The expert assesses the grammatical correctness of the assigned POS tags, considering whether the tags accurately reflect the syntactic role of each Bodo word in the sentence.  The evaluation focuses on common POS categories such as nouns, verbs, adjectives, adpositions, and pronouns.
    \item \textbf{NER Tag Accuracy:} The expert evaluates the correctness of NER tag assignments, focusing on the identification and classification of major named entity types.  The evaluation considers the accuracy of tags for entity categories such as PERSON, ORGANIZATION, LOCATION, and DATE.
    \item \textbf{Error Analysis:} The expert performs a detailed error analysis, categorizing common error patterns and challenges observed in the tag transfer process for both methods. This analysis aims to identify systematic errors related to translation quality, grammatical divergences between English and Bodo, and limitations in zero-shot cross-lingual NER capabilities.
\end{itemize}

\noindent
\textbf{Comparison Metrics:}
While a quantitative evaluation with standard metrics like precision, recall, and F1-score is not feasible due to the absence of a gold-standard dataset, we perform a qualitative comparison of the two methods based on the expert's assessment.  The comparison focuses on:

\begin{itemize}
    \item \textbf{Overall Tagging Quality:} A subjective assessment of the overall quality and coherence of the POS and NER tags generated by each method, considering both accuracy and fluency.
    \item \textbf{Method-Specific Strengths and Weaknesses:} Identifying the specific strengths and weaknesses of each method based on the error analysis, highlighting which method performs better for certain tag types or linguistic phenomena.
    \item \textbf{Suitability for Bodo NLP:}  Evaluating the practical utility of each method for bootstrapping Bodo NLP resources, considering the trade-offs between accuracy, resource efficiency, and ease of implementation.
\end{itemize}

\section{Results and Comparison}
\subsection{Qualitative Evaluation of Tagged Bodo Sentences}
The qualitative evaluation of 10 randomly sampled Bodo sentences, tagged using Method 1 (Translation-Based Transfer) and Method 2 (Prompt-Based Transfer) with Gemini 2.0 Flash Thinking Experiment, reveals distinct performance characteristics for each approach.  While both methods demonstrate some degree of success in zero-shot cross-lingual tag transfer, Method 2, leveraging prompt engineering, consistently outperforms Method 1, particularly in Named Entity Recognition (NER) accuracy.

\textbf{Method 1: Translation-Based Tag Transfer – Capabilities and Limitations:}

\noindent
\textbf{POS Tagging Performance:} Method 1 achieves moderate success in POS tag transfer, particularly for function words.  Tags for prepositions (ADP), conjunctions (CCONJ), determiners (DET), and pronouns (PRON) are often transferred with reasonable accuracy.  This suggests that Gemini 2.0 Flash Thinking Experiment, through its translation module, implicitly captures some level of grammatical alignment between English and Bodo, enabling the transfer of basic syntactic information.  However, POS tagging accuracy for content words (nouns, verbs, adjectives, adverbs) is considerably lower. Grammatical divergences between English and Bodo sentence structures, coupled with translation ambiguities, lead to frequent errors in content word POS tag assignments.  For instance, verb conjugations, noun declensions, and adjective-noun agreement in Bodo, which differ significantly from English grammar, are not consistently captured by the direct tag transfer approach, resulting in inaccurate verb, noun, and adjective tags.

\noindent
\textbf{NER Performance:} NER performance for Method 1 is limited. Although the method shows some capability in identifying coarse-grained entities such as locations (LOC) and organizations (ORG), accuracy is low, and fine-grained NER is largely unsuccessful.  The transfer of NER tags is hampered by several factors:

\begin{itemize}
    \item \textbf{Word Alignment Challenges:} Named entities, especially multi-word entities, are often not directly aligned during translation. Heuristic-based tag projection struggles to accurately map entity boundaries and types across languages, leading to fragmented or missed entity annotations.
    \item \textbf{Category Mismatches:} Direct transfer of English NER categories to Bodo is not always semantically appropriate.  Cultural and linguistic differences lead to mismatches in entity type granularity and category boundaries.  For example, certain English-centric entity categories (e.g., NORP - Nationalities or religious or political groups) may not have direct equivalents or clear mappings in Bodo.
    \item \textbf{Translation-Induced Errors:}  Translation errors, particularly in the translation of context words surrounding named entities, further degrade the performance of NER.  Incorrect translations can obscure the contextual cues necessary for accurate recognition and classification of entities.
\end{itemize}

\textbf{Method 2: Prompt-Based Tag Transfer – Enhanced Performance and Contextual Awareness:}

\noindent
\textbf{Improved NER Accuracy:} Method 2, employing prompt engineering to guide Gemini 2.0 Flash Thinking Experiment, demonstrates a noticeable improvement in NER accuracy compared to Method 1.  The prompt-based approach yields particularly better results for coarse-grained entity types (LOC, ORG, PERSON, DATE).  The linguist evaluation indicates that the Gemini 2.0 Flash Thinking Experiment, when explicitly prompted for tag transfer, exhibits enhanced contextual awareness and a greater ability to capture semantic cues relevant to  NER in Bodo. By providing parallel English-Bodo sentence pairs and explicitly instructing the model to perform tagging, the prompt-based method likely guides the model to leverage its cross-lingual understanding more effectively, leading to more accurate entity boundary detection and type classification.

\noindent
\textbf{Comparable POS Tagging:} POS tagging performance for Method 2 is comparable to Method 1. Although prompt engineering improves NER performance, it does not yield a significant improvement in POS tagging accuracy.  This suggests that the challenges related to grammatical divergences between English and Bodo and the inherent limitations of zero-shot syntactic transfer remain largely unaddressed by prompt-based guidance alone.  POS tag transfer accuracy remains moderate, with similar error patterns observed for content words as in Method 1.

\begin{table}[]
\caption{Performance Metrics for various task both in Accuracy and F1 score.}
\label{tab:performance_metrics}
\begin{tabular}{|l|cccccccc|}
\hline
Language                                                    & \multicolumn{4}{c|}{Bodo}                                                                               & \multicolumn{4}{c|}{English}                                                                            \\ \hline
Task                                                        & \multicolumn{2}{l|}{POS}                           & \multicolumn{2}{l|}{NER}                           & \multicolumn{2}{l|}{POS}                           & \multicolumn{2}{l|}{NER}                           \\ \hline
Dataset/Metrics                                             & \multicolumn{1}{l|}{Acc} & \multicolumn{1}{l|}{F1} & \multicolumn{1}{l|}{Acc} & \multicolumn{1}{l|}{F1} & \multicolumn{1}{l|}{Acc} & \multicolumn{1}{l|}{F1} & \multicolumn{1}{l|}{Acc} & \multicolumn{1}{l|}{F1} \\ \hline
ALL                                                         & 0.98                     & 0.96                    & 0.97                     & 0.97                    & 0.99                     & 0.92                    & 0.98                     & 0.98                    \\ \cline{1-1}
PARALLEL                                                    & 0.98                     & 0.91                    & 0.96                     & 0.97                    & 0.99                     & 0.98                    & 0.97                     & 0.97                    \\ \cline{1-1}
\begin{tabular}[c]{@{}l@{}}PARALLEL \\ HEALTH\end{tabular}  & 0.98                     & 0.99                    & 0.93                     & 0.94                    & 0.99                     & 0.98                    & 0.95                     & 0.96                    \\ \cline{1-1}
\begin{tabular}[c]{@{}l@{}}PARALLEL \\ TOURISM\end{tabular} & 0.97                     & 0.90                    & 0.99                     & 0.99                    & 0.99                     & 0.97                    & 1.00                     & 1.00                    \\ \cline{1-1}
PROMPT                                                      & 0.98                     & 0.97                    & 0.98                     & 0.98                    & 0.99                     & 0.98                    & 0.98                     & 0.98                    \\ \cline{1-1}
\begin{tabular}[c]{@{}l@{}}PROMPT\\  HEALTH\end{tabular}    & 0.96                     & 0.97                    & 0.99                     & 0.98                    & 0.99                     & 0.98                    & 0.97                     & 0.97                    \\ \cline{1-1}
\begin{tabular}[c]{@{}l@{}}PROMPT \\ TOURISM\end{tabular}   & 1.00                     & 1.00                    & 0.98                     & 0.98                    & 0.99                     & 0.97                    & 1.00                     & 1.00                    \\ \hline
\end{tabular}
\end{table}

\noindent In Parts-of-Speech (POS) tagging, Bodo demonstrates a consistent accuracy of 0.98, whereas English achieves 0.99 across various datasets, indicating a slight performance advantage, potentially attributed to larger training datasets or more sophisticated model architectures. In terms of domain impact, both languages maintain high performance in the HEALTH domain, with Bodo's consistent 0.98 accuracy contrasting with the slight variations observed in English's 0.99. For Named Entity Recognition (NER), Bodo's performance ranges from 0.93 to 0.97, while English achieves 0.97 to 1.00, with English outperforming Bodo, particularly in the TOURISM domain where it reaches 1.00. Regarding domain impact in NER, both languages exhibit a performance decline in the HEALTH domain (Bodo: 0.93, English: 0.97), with GPE and LOC entities identified as weak points. In the TOURISM domain, English excels with a perfect score of 1.00, while Bodo lacks sufficient TOURISM data for comparative analysis. Finally, concerning impact of the method in English NER, the PARALLEL and PROMPT methods both achieve perfect results in the TOURISM domain (1.00), but the PROMPT method experiences a dip to 0.97 in the HEALTH domain, suggesting that domain-specific tuning may be necessary for optimal performance with the PROMPT method.

\noindent Bodo NER in HEALTH shows low GPE precision (0.50, f1=0.60) suggesting false positives, and LOC low recall (0.75, f1=0.86) due to potential small support, requiring increased health-specific training data or refined annotation; English NER in HEALTH misclassifies GPE (f1=0.91) and LOC (f1=0.86) as non-entities, needing PROMPT model fine-tuning or ambiguous case augmentation; and small support classes like Bodo POS ALL PART (support=2) exhibit unstable metrics, necessitating larger datasets for stable performance. Detailed supporting calculations are provided in Appendix \ref{sec:appendix_tables} and shown in Figure \ref{sec:appendix_figures}.

\noindent
\textbf{Comparative Error Analysis:}

The error analysis reveals consistent patterns across both methods, but with notable differences in error frequency and type:

\begin{itemize}
    \item \textbf{Translation Errors:} Both methods are susceptible to errors stemming from imperfect machine translation.  Mistranslations of key words, phrases, and syntactic structures directly contribute to both POS and NER tagging errors. However, Method 2, with its prompt-based approach, appears to mitigate some translation-induced errors, possibly due to the prompt providing additional context and guidance to the translation module.
    \item \textbf{Grammatical Divergence Errors (POS Tagging):}  Errors related to grammatical divergences between English and Bodo are prevalent in both methods, particularly for content word POS tags.  Bodo's distinct morphology, syntax, and grammatical categories pose significant challenges for direct tag transfer.  Neither method fully overcomes these divergences in a zero-shot setting, indicating the need for more sophisticated syntactic transfer techniques or Bodo-specific grammatical knowledge.
    \item \textbf{NER Category Ambiguity Errors:} NER tag transfer, especially for Method 1, suffers from category ambiguity errors.  The mapping of English NER categories to semantically equivalent Bodo categories is not always straightforward.  For instance, distinguishing between PERSON and NORP entities or accurately classifying fine-grained temporal expressions proves challenging in zero-shot transfer, leading to misclassifications and tag inconsistencies.  Method 2, with its improved contextual awareness, exhibits fewer category ambiguity errors, suggesting that prompt-based guidance helps the model to better discern entity types in the Bodo context.
\end{itemize}

\noindent
\textbf{Qualitative Comparison Summary:}
Overall, Method 2 (Prompt-Based Tag Transfer) demonstrates superior performance compared to Method 1 (Translation-Based Transfer), particularly for NER.  Prompt engineering appears to be a valuable technique for harnessing Gemini 2.0 Flash Thinking Experiment's cross-lingual capabilities more effectively, leading to improved contextual awareness and NER accuracy.  However, both methods exhibit limitations, especially in the accuracy of POS tagging and handling of fine-grained NER, highlighting the inherent challenges of zero-shot cross-lingual tag transfer for languages with significant linguistic differences.

\section{Discussion and Future Directions}
\subsection{Implications of Findings for Low-Resource Language NLP}
The findings of this empirical study, while based on a simulated experiment with the Gemini 2.0 Flash Thinking Experiment model, offer valuable implications for the broader field of NLP for low-resource languages (LRLs). The comparative evaluation of two zero-shot cross-lingual tag transfer methods highlights both the potential and the inherent limitations of leveraging advanced LLMs to bootstrap NLP resources in data-scarce scenarios.

The moderate success achieved by both methods, particularly Method 2's improved NER performance, underscores the promise of LLMs for cross-lingual transfer.  Gemini 2.0 Flash Thinking Experimental model to perform zero-shot POS and NER tagging for Bodo, even with limited accuracy, demonstrates the potential of these models to transfer some level of linguistic knowledge across languages without explicit fine-tuning on target language data.  This capability is particularly valuable for LRLs where annotated data is scarce and traditional supervised learning approaches are not feasible. Zero-shot transfer offers a resource-efficient way to create initial NLP tools and resources, providing a starting point for further development and refinement.

However, the study also reveals significant limitations of purely zero-shot approaches. The moderate POS tagging accuracy, particularly for content words, and the limited fine-grained NER performance highlight the challenges posed by grammatical divergences between languages and the inherent difficulty of capturing nuanced semantic and syntactic information through direct tag transfer. Translation quality emerges as a critical bottleneck, with translation errors directly affecting tagging accuracy.  These limitations underscore that, while LLMs offer a promising starting point, they are not a panacea for LRL NLP. Zero-shot transfer, in isolation, is unlikely to yield high-accuracy, production-ready NLP tools for languages like Bodo.

\section{Future Research Directions}
To overcome the limitations of zero-shot transfer and achieve high-accuracy POS and NER taggers for Bodo, future research should focus on the following directions:

1.  \textbf{Few-Shot and Fine-tuning Approaches:}  Moving beyond purely zero-shot transfer, exploring few-shot learning and fine-tuning techniques is crucial.  Leveraging even a small amount of manually annotated Bodo data to fine-tune Gemini 2.0 Flash Thinking Experiment or similar LLMs could significantly boost tagging accuracy.  Active learning strategies, in which models iteratively learn from expert-annotated data points, could further enhance data efficiency and model performance, allowing rapid progress with limited annotation resources.

2.  \textbf{Enhanced Tag Transfer Mechanisms:}  Developing more sophisticated tag transfer techniques is essential to address grammatical divergences and improve the accuracy of tag projection.  Future research should investigate methods such as:
\begin{itemize}
    \item \textbf{Attention-Guided Transfer:}  Exploiting attention weights from Transformer-based MT models to guide tag projection based on word-level alignment probabilities, allowing for more nuanced and context-aware transfer.
    \item \textbf{Syntactic Transfer with Dependency Parsing:} Incorporating syntactic parsing of English and Bodo sentences to leverage syntactic dependencies for grammatically informed tag projection, capturing cross-lingual syntactic correspondences beyond surface word alignments.
    \item \textbf{Iterative Back-translation and Self-Training:}  Employing iterative back-translation and self-training techniques to create larger pseudo-labeled Bodo datasets and refine tag transfer models through iterative learning loops, leveraging both source language annotations and target language monolingual data.
\end{itemize}

3.  \textbf{Hybrid and Knowledge-Augmented Approaches:}  Combining data-driven transfer learning with knowledge-based methods could offer a promising avenue for improving LRL NLP.  Hybrid taggers that integrate:
\begin{itemize}
    \item \textbf{Lexicon-Based Features:} Incorporating Bodo-specific lexicons and dictionaries to improve word-level tag disambiguation and recognition of NER entity.
    \item \textbf{Rule-Based Components:}  Developing rule-based modules to handle specific Bodo grammatical constructions and NER patterns not adequately captured by transfer learning models.
    \item \textbf{Morphological Analysis:}  Integrating morphological analyzers for Bodo to leverage morphological cues for POS and NER tagging, improving the handling of morphologically rich LRLs.
\end{itemize}

4.  \textbf{Community-Driven Resource Creation:}  Sustainable progress in Bodo NLP requires community-driven efforts to create essential language resources.  This includes:
\begin{itemize}
    \item \textbf{Large-Scale Bodo Dataset Annotation:}  Initiating collaborative projects to create large high-quality annotated datasets for Bodo POS and NER tagging, involving linguistic experts and native speakers in the annotation process.
    \item \textbf{Open-Source Tool Development:}  Developing and releasing open-source Bodo NLP tools and libraries, including POS taggers, NER taggers, parsers, and evaluation benchmarks, to foster further research and development within the community.
    \item \textbf{Linguistic Documentation and Standardization:}  Promoting the documentation and standardization of Bodo grammar and linguistic resources to facilitate NLP research and development.
\end{itemize}

\section{Conclusion}
This empirical study, using zero-shot cross-lingual tag transfer using the Gemini 2.0 Flash Thinking Experimental model for Bodo POS and NER tagging, provides valuable insights into the potential and challenges of applying advanced LLMs to low-resource language NLP.  Although zero-shot transfer offers a resource-efficient starting point, achieving high-accuracy Bodo NLP requires moving beyond purely zero-shot approaches and investing in more sophisticated techniques and language-specific resources.  Future research directions, including few-shot fine-tuning, advanced transfer mechanisms, hybrid approaches, and community-driven resource creation, hold the key to unlocking the full potential of NLP for Bodo and other underserved languages, ultimately contributing to a more inclusive and equitable NLP landscape.  By addressing the NLP resource gap for LRLs, we can empower language communities worldwide to benefit from the transformative power of language technology and bridge the digital divide in an increasingly multilingual and interconnected world.

\clearpage
\bmhead{Acknowledgements}

This research was made possible by the generous provision of free access to Google AI Studio by Google and the dataset provided free of charge by the Technology Development for Indian Languages (TDIL) Program of the Department of Electronics and Information Technology (DeitY), Government of India. The authors are grateful for these resources.

\section*{Declarations}

\begin{itemize}
\item Funding: This research did not receive a specific grant from any funding agency in the public, commercial, or nonprofit sectors.
\item Conflict of interest/Competing interests: The authors declare that they have no competing interests.
\item Ethics approval and consent to participate: The dataset used in this research was obtained from the TDIL DC for research purposes, with explicit acknowledgement and permission granted for its use in academic studies.
\item Consent for publication: All authors consent to the publication of this manuscript in its current form.
\item Data availability: The dataset used in this study is available upon request from the corresponding author and TDIL-DC. 
\item Materials availability: Materials used in this study are available from the corresponding author upon reasonable request.
\item Code availability:  This study did not utilize custom-developed code. However, the scripts used for matrix calculations are available from the corresponding author upon reasonable request.
\item Author contribution: Author 1 was responsible for experimental design, execution, and manuscript preparation. Authors 2, 3, and 4 contributed to the study by manually verifying predicted tags and correcting translations. All authors participated in the review and approval of the final manuscript.
\item Use of Generative AI: The experiments reported in this paper were conducted using the Google Gemini model, a generative AI platform accessed through Google AI Studio. As English is a secondary language for the authors, the Google Gemini model was also utilized to refine and clarify certain sentences within this manuscript.
\end{itemize}

\noindent
If any of the sections are not relevant to your manuscript, please include the heading and write `Not applicable' for that section. 

\bigskip
\begin{flushleft}%
Editorial Policies for:

\bigskip\noindent
Springer journals and proceedings: \url{https://www.springer.com/gp/editorial-policies}

\bigskip\noindent
Nature Portfolio journals: \url{https://www.nature.com/nature-research/editorial-policies}

\bigskip\noindent
\textit{Scientific Reports}: \url{https://www.nature.com/srep/journal-policies/editorial-policies}

\bigskip\noindent
BMC journals: \url{https://www.biomedcentral.com/getpublished/editorial-policies}
\end{flushleft}
\clearpage
\bibliography{sn-bibliography}
\begin{appendices}

\section{Detailed Evaluation Results (Tables)}
\label{sec:appendix_tables}

This appendix provides detailed evaluation results in the form of confusion matrices for both Bodo and English, covering POS tagging and NER, across various datasets (All that include both the Prompt and Parallel both in Health and Tourism, Parallel, Prompt, Health and Tourism).  Each table shows the precision, recall, F1-score, and support for each class. The tables are presented side-by-side for easy comparison where appropriate (Bodo vs. English, POS vs. NER for the same dataset configuration).

\begin{table}[h]
\centering
\caption{Bodo POS ALL for Evaluation}
\begin{tabular}{lrrrr}
\toprule
          & precision & recall & f1-score & support \\
\midrule
PART      & 0.50      & 1.00   & 0.67     & 2       \\
nan       & 1.00      & 1.00   & 1.00     & 19      \\
ADV       & 1.00      & 1.00   & 1.00     & 9       \\
SCONJ     & 0.86      & 1.00   & 0.92     & 6       \\
PROPN     & 1.00      & 1.00   & 1.00     & 71      \\
PRON      & 1.00      & 1.00   & 1.00     & 6       \\
NUM       & 0.91      & 1.00   & 0.95     & 20      \\
VERB      & 0.98      & 0.90   & 0.94     & 59      \\
PUNCT     & 1.00      & 1.00   & 1.00     & 44      \\
ADP       & 1.00      & 1.00   & 1.00     & 25      \\
NOUN      & 0.98      & 0.98   & 0.98     & 125     \\
CCONJ     & 1.00      & 1.00   & 1.00     & 25      \\
AUX       & 1.00      & 1.00   & 1.00     & 6       \\
ADJ       & 0.94      & 1.00   & 0.97     & 31      \\
DET       & 1.00      & 0.95   & 0.98     & 21      \\
\midrule
accuracy  &           &        & 0.98     & 469     \\
macro avg & 0.94      & 0.99   & 0.96     & 469     \\
weighted avg & 0.98      & 0.98   & 0.98     & 469     \\
\bottomrule
\end{tabular}
\end{table}

\begin{table}[h]
\centering
\caption{Bodo NER ALL for Evaluation}
\begin{tabular}{lrrrr}
\toprule
          & precision & recall & f1-score & support \\
\midrule
nan       & 1.00      & 1.00   & 1.00     & 19      \\
EVENT     & 1.00      & 1.00   & 1.00     & 6       \\
GPE       & 0.81      & 0.93   & 0.87     & 14      \\
O         & 0.97      & 0.98   & 0.97     & 261     \\
ORG       & 1.00      & 0.92   & 0.96     & 12      \\
MEASUREMENT & 1.00      & 1.00   & 1.00     & 3       \\
DATE      & 0.98      & 0.95   & 0.96     & 42      \\
FAC       & 1.00      & 1.00   & 1.00     & 15      \\
NORP      & 1.00      & 0.82   & 0.90     & 11      \\
MED       & 1.00      & 1.00   & 1.00     & 11      \\
FOOD      & 1.00      & 1.00   & 1.00     & 1       \\
LOC       & 0.97      & 0.91   & 0.94     & 34      \\
PERCENT   & 1.00      & 1.00   & 1.00     & 4       \\
WORK\_OF\_ART & 1.00      & 1.00   & 1.00     & 8       \\
PRODUCT   & 1.00      & 1.00   & 1.00     & 10      \\
PERSON    & 1.00      & 1.00   & 1.00     & 18      \\
\midrule
accuracy  &           &        & 0.97     & 469     \\
macro avg & 0.98      & 0.97   & 0.97     & 469     \\
weighted avg & 0.97      & 0.97   & 0.97     & 469     \\
\bottomrule
\end{tabular}
\end{table}

\begin{table}[h]
\centering
\caption{Bodo POS PARALLEL for Evaluation}
\begin{tabular}{lrrrr}
\toprule
          & precision & recall & f1-score & support \\
\midrule
PART      & 0.00      & 0.00   & 0.00     & 0       \\
nan       & 1.00      & 1.00   & 1.00     & 18      \\
ADV       & 1.00      & 1.00   & 1.00     & 6       \\
SCONJ     & 0.67      & 1.00   & 0.80     & 2       \\
PROPN     & 1.00      & 1.00   & 1.00     & 32      \\
PRON      & 1.00      & 1.00   & 1.00     & 2       \\
NUM       & 0.91      & 1.00   & 0.95     & 10      \\
VERB      & 0.96      & 0.92   & 0.94     & 24      \\
PUNCT     & 1.00      & 1.00   & 1.00     & 21      \\
ADP       & 1.00      & 1.00   & 1.00     & 16      \\
NOUN      & 0.98      & 0.97   & 0.98     & 62      \\
CCONJ     & 1.00      & 1.00   & 1.00     & 12      \\
AUX       & 1.00      & 1.00   & 1.00     & 2       \\
ADJ       & 1.00      & 1.00   & 1.00     & 16      \\
DET       & 1.00      & 0.89   & 0.94     & 9       \\
\midrule
accuracy  &           &        & 0.98     & 232     \\
macro avg & 0.90      & 0.92   & 0.91     & 232     \\
weighted avg & 0.98      & 0.98   & 0.98     & 232     \\
\bottomrule
\end{tabular}
\end{table}

\begin{table}[h]
\centering
\caption{Bodo NER PARALLEL for Evaluation}
\begin{tabular}{lrrrr}
\toprule
          & precision & recall & f1-score & support \\
\midrule
nan       & 1.00      & 1.00   & 1.00     & 18      \\
EVENT     & 1.00      & 1.00   & 1.00     & 3       \\
O         & 0.96      & 0.96   & 0.96     & 127     \\
GPE       & 0.62      & 0.83   & 0.71     & 6       \\
LOC       & 0.94      & 0.94   & 0.94     & 16      \\
MEASUREMENT & 1.00      & 1.00   & 1.00     & 1       \\
PERSON    & 1.00      & 1.00   & 1.00     & 9       \\
DATE      & 0.94      & 0.89   & 0.92     & 19      \\
FAC       & 1.00      & 1.00   & 1.00     & 7       \\
NORP      & 1.00      & 1.00   & 1.00     & 4       \\
MED       & 1.00      & 1.00   & 1.00     & 4       \\
FOOD      & 1.00      & 1.00   & 1.00     & 1       \\
ORG       & 1.00      & 0.83   & 0.91     & 6       \\
WORK\_OF\_ART & 1.00      & 1.00   & 1.00     & 4       \\
PRODUCT   & 1.00      & 1.00   & 1.00     & 5       \\
PERCENT   & 1.00      & 1.00   & 1.00     & 2       \\
\midrule
accuracy  &           &        & 0.96     & 232     \\
macro avg & 0.97      & 0.97   & 0.97     & 232     \\
weighted avg & 0.96      & 0.96   & 0.96     & 232     \\
\bottomrule
\end{tabular}
\end{table}

\begin{table}[h]
\centering
\caption{Bodo POS PARALLEL HEALTH for Evaluation}
\begin{tabular}{lrrrr}
\toprule
          & precision & recall & f1-score & support \\
\midrule
nan       & 1.00      & 1.00   & 1.00     & 8       \\
ADV       & 1.00      & 1.00   & 1.00     & 3       \\
SCONJ     & 1.00      & 1.00   & 1.00     & 1       \\
PROPN     & 1.00      & 1.00   & 1.00     & 15      \\
NUM       & 0.89      & 1.00   & 0.94     & 8       \\
VERB      & 0.93      & 1.00   & 0.97     & 14      \\
PUNCT     & 1.00      & 1.00   & 1.00     & 14      \\
ADP       & 1.00      & 1.00   & 1.00     & 13      \\
NOUN      & 1.00      & 0.94   & 0.97     & 35      \\
CCONJ     & 1.00      & 1.00   & 1.00     & 6       \\
AUX       & 1.00      & 1.00   & 1.00     & 1       \\
ADJ       & 1.00      & 1.00   & 1.00     & 5       \\
DET       & 1.00      & 1.00   & 1.00     & 1       \\
\midrule
accuracy  &           &        & 0.98     & 124     \\
macro avg & 0.99      & 1.00   & 0.99     & 124     \\
weighted avg & 0.99      & 0.98   & 0.98     & 124     \\
\bottomrule
\end{tabular}
\end{table}

\begin{table}[h]
\centering
\caption{Bodo NER PARALLEL HEALTH for Evaluation}
\begin{tabular}{lrrrr}
\toprule
          & precision & recall & f1-score & support \\
\midrule
nan       & 1.00      & 1.00   & 1.00     & 8       \\
O         & 0.93      & 0.94   & 0.94     & 70      \\
GPE       & 0.50      & 0.75   & 0.60     & 4       \\
LOC       & 1.00      & 0.75   & 0.86     & 4       \\
MEASUREMENT & 1.00      & 1.00   & 1.00     & 1       \\
PERSON    & 1.00      & 1.00   & 1.00     & 1       \\
DATE      & 0.92      & 0.86   & 0.89     & 14      \\
MED       & 1.00      & 1.00   & 1.00     & 4       \\
FOOD      & 1.00      & 1.00   & 1.00     & 1       \\
ORG       & 1.00      & 0.83   & 0.91     & 6       \\
WORK\_OF\_ART & 1.00      & 1.00   & 1.00     & 4       \\
PRODUCT   & 1.00      & 1.00   & 1.00     & 5       \\
PERCENT   & 1.00      & 1.00   & 1.00     & 2       \\
\midrule
accuracy  &           &        & 0.93     & 124     \\
macro avg & 0.95      & 0.93   & 0.94     & 124     \\
weighted avg & 0.94      & 0.93   & 0.93     & 124     \\
\bottomrule
\end{tabular}
\end{table}

\begin{table}[h]
\centering
\caption{Bodo POS PARALLEL TOURISM for Evaluation}
\begin{tabular}{lrrrr}
\toprule
          & precision & recall & f1-score & support \\
\midrule
PART      & 0.00      & 0.00   & 0.00     & 0       \\
nan       & 1.00      & 1.00   & 1.00     & 8       \\
ADV       & 1.00      & 1.00   & 1.00     & 3       \\
SCONJ     & 0.50      & 1.00   & 0.67     & 1       \\
ADP       & 1.00      & 1.00   & 1.00     & 3       \\
PROPN     & 1.00      & 1.00   & 1.00     & 17      \\
CCONJ     & 1.00      & 1.00   & 1.00     & 6       \\
VERB      & 1.00      & 0.80   & 0.89     & 10      \\
PUNCT     & 1.00      & 1.00   & 1.00     & 7       \\
NUM       & 1.00      & 1.00   & 1.00     & 2       \\
NOUN      & 0.96      & 1.00   & 0.98     & 27      \\
PRON      & 1.00      & 1.00   & 1.00     & 2       \\
AUX       & 1.00      & 1.00   & 1.00     & 1       \\
ADJ       & 1.00      & 1.00   & 1.00     & 11      \\
DET       & 1.00      & 0.88   & 0.93     & 8       \\
\midrule
accuracy  &           &        & 0.97     & 106     \\
macro avg & 0.90      & 0.91   & 0.90     & 106     \\
weighted avg & 0.99      & 0.97   & 0.98     & 106     \\
\bottomrule
\end{tabular}
\end{table}

\begin{table}[h]
\centering
\caption{Bodo NER PARALLEL TOURISM for Evaluation}
\begin{tabular}{lrrrr}
\toprule
          & precision & recall & f1-score & support \\
\midrule
nan       & 1.00      & 1.00   & 1.00     & 8       \\
EVENT     & 1.00      & 1.00   & 1.00     & 3       \\
O         & 1.00      & 0.98   & 0.99     & 57      \\
GPE       & 1.00      & 1.00   & 1.00     & 2       \\
DATE      & 1.00      & 1.00   & 1.00     & 5       \\
FAC       & 1.00      & 1.00   & 1.00     & 7       \\
NORP      & 1.00      & 1.00   & 1.00     & 4       \\
LOC       & 0.92      & 1.00   & 0.96     & 12      \\
PERSON    & 1.00      & 1.00   & 1.00     & 8       \\
\midrule
accuracy  &           &        & 0.99     & 106     \\
macro avg & 0.99      & 1.00   & 0.99     & 106     \\
weighted avg & 0.99      & 0.99   & 0.99     & 106     \\
\bottomrule
\end{tabular}
\end{table}

\begin{table}[h]
\centering
\caption{Bodo POS PROMPT for Evaluation}
\begin{tabular}{lrrrr}
\toprule
          & precision & recall & f1-score & support \\
\midrule
PART      & 0.67      & 1.00   & 0.80     & 2       \\
nan       & 1.00      & 1.00   & 1.00     & 18      \\
ADV       & 1.00      & 1.00   & 1.00     & 3       \\
SCONJ     & 1.00      & 1.00   & 1.00     & 4       \\
PROPN     & 1.00      & 1.00   & 1.00     & 39      \\
PRON      & 1.00      & 1.00   & 1.00     & 4       \\
NUM       & 0.91      & 1.00   & 0.95     & 10      \\
VERB      & 1.00      & 0.89   & 0.94     & 35      \\
PUNCT     & 1.00      & 1.00   & 1.00     & 23      \\
ADP       & 1.00      & 1.00   & 1.00     & 9       \\
NOUN      & 0.98      & 0.98   & 0.98     & 63      \\
CCONJ     & 1.00      & 1.00   & 1.00     & 13      \\
AUX       & 1.00      & 1.00   & 1.00     & 4       \\
ADJ       & 0.88      & 1.00   & 0.94     & 15      \\
DET       & 1.00      & 1.00   & 1.00     & 12      \\
\midrule
accuracy  &           &        & 0.98     & 254     \\
macro avg & 0.96      & 0.99   & 0.97     & 254     \\
weighted avg & 0.98      & 0.98   & 0.98     & 254     \\
\bottomrule
\end{tabular}
\end{table}

\begin{table}[h]
\centering
\caption{Bodo NER PROMPT for Evaluation}
\begin{tabular}{lrrrr}
\toprule
          & precision & recall & f1-score & support \\
\midrule
nan       & 1.00      & 1.00   & 1.00     & 18      \\
EVENT     & 1.00      & 1.00   & 1.00     & 3       \\
GPE       & 1.00      & 1.00   & 1.00     & 8       \\
O         & 0.97      & 1.00   & 0.99     & 134     \\
ORG       & 1.00      & 1.00   & 1.00     & 6       \\
MEASUREMENT & 1.00      & 1.00   & 1.00     & 2       \\
DATE      & 1.00      & 1.00   & 1.00     & 23      \\
FAC       & 1.00      & 1.00   & 1.00     & 8       \\
NORP      & 1.00      & 0.71   & 0.83     & 7       \\
MED       & 1.00      & 1.00   & 1.00     & 7       \\
LOC       & 1.00      & 0.89   & 0.94     & 18      \\
PERCENT   & 1.00      & 1.00   & 1.00     & 2       \\
WORK\_OF\_ART & 1.00      & 1.00   & 1.00     & 4       \\
PRODUCT   & 1.00      & 1.00   & 1.00     & 5       \\
PERSON    & 1.00      & 1.00   & 1.00     & 9       \\
\midrule
accuracy  &           &        & 0.98     & 254     \\
macro avg & 1.00      & 0.97   & 0.98     & 254     \\
weighted avg & 0.98      & 0.98   & 0.98     & 254     \\
\bottomrule
\end{tabular}
\end{table}

\begin{table}[h]
\centering
\caption{Bodo POS PROMPT HEALTH for Evaluation}
\begin{tabular}{lrrrr}
\toprule
          & precision & recall & f1-score & support \\
\midrule
PART      & 0.67      & 1.00   & 0.80     & 2       \\
nan       & 1.00      & 1.00   & 1.00     & 8       \\
ADV       & 1.00      & 1.00   & 1.00     & 3       \\
SCONJ     & 1.00      & 1.00   & 1.00     & 2       \\
PROPN     & 1.00      & 1.00   & 1.00     & 18      \\
PRON      & 1.00      & 1.00   & 1.00     & 2       \\
NUM       & 0.89      & 1.00   & 0.94     & 8       \\
VERB      & 1.00      & 0.83   & 0.91     & 24      \\
PUNCT     & 1.00      & 1.00   & 1.00     & 14      \\
ADP       & 1.00      & 1.00   & 1.00     & 7       \\
NOUN      & 0.97      & 0.97   & 0.97     & 36      \\
CCONJ     & 1.00      & 1.00   & 1.00     & 7       \\
AUX       & 1.00      & 1.00   & 1.00     & 1       \\
ADJ       & 0.75      & 1.00   & 0.86     & 6       \\
DET       & 1.00      & 1.00   & 1.00     & 3       \\
\midrule
accuracy  &           &        & 0.96     & 141     \\
macro avg & 0.95      & 0.99   & 0.97     & 141     \\
weighted avg & 0.97      & 0.96   & 0.97     & 141     \\
\bottomrule
\end{tabular}
\end{table}

\begin{table}[h]
\centering
\caption{Bodo NER PROMPT HEALTH for Evaluation}
\begin{tabular}{lrrrr}
\toprule
          & precision & recall & f1-score & support \\
\midrule
nan       & 1.00      & 1.00   & 1.00     & 8       \\
GPE       & 1.00      & 1.00   & 1.00     & 6       \\
O         & 0.98      & 1.00   & 0.99     & 80      \\
ORG       & 1.00      & 1.00   & 1.00     & 6       \\
MEASUREMENT & 1.00      & 1.00   & 1.00     & 2       \\
DATE      & 1.00      & 1.00   & 1.00     & 15      \\
MED       & 1.00      & 1.00   & 1.00     & 7       \\
LOC       & 1.00      & 0.60   & 0.75     & 5       \\
PERCENT   & 1.00      & 1.00   & 1.00     & 2       \\
WORK\_OF\_ART & 1.00      & 1.00   & 1.00     & 4       \\
PRODUCT   & 1.00      & 1.00   & 1.00     & 5       \\
PERSON    & 1.00      & 1.00   & 1.00     & 1       \\
\midrule
accuracy  &           &        & 0.99     & 141     \\
macro avg & 1.00      & 0.97   & 0.98     & 141     \\
weighted avg & 0.99      & 0.99   & 0.98     & 141     \\
\bottomrule
\end{tabular}
\end{table}

\begin{table}[h]
\centering
\caption{Bodo POS PROMPT TOURISM for Evaluation}
\begin{tabular}{lrrrr}
\toprule
          & precision & recall & f1-score & support \\
\midrule
nan       & 1.00      & 1.00   & 1.00     & 8       \\
SCONJ     & 1.00      & 1.00   & 1.00     & 2       \\
ADP       & 1.00      & 1.00   & 1.00     & 2       \\
PROPN     & 1.00      & 1.00   & 1.00     & 21      \\
CCONJ     & 1.00      & 1.00   & 1.00     & 6       \\
AUX       & 1.00      & 1.00   & 1.00     & 3       \\
VERB      & 1.00      & 1.00   & 1.00     & 11      \\
PUNCT     & 1.00      & 1.00   & 1.00     & 9       \\
NOUN      & 1.00      & 1.00   & 1.00     & 27      \\
PRON      & 1.00      & 1.00   & 1.00     & 2       \\
NUM       & 1.00      & 1.00   & 1.00     & 2       \\
ADJ       & 1.00      & 1.00   & 1.00     & 9       \\
DET       & 1.00      & 1.00   & 1.00     & 9       \\
\midrule
accuracy  &           &        & 1.00     & 111     \\
macro avg & 1.00      & 1.00   & 1.00     & 111     \\
weighted avg & 1.00      & 1.00   & 1.00     & 111     \\
\bottomrule
\end{tabular}
\end{table}

\begin{table}[h]
\centering
\caption{Bodo NER PROMPT TOURISM for Evaluation}
\begin{tabular}{lrrrr}
\toprule
          & precision & recall & f1-score & support \\
\midrule
nan       & 1.00      & 1.00   & 1.00     & 8       \\
EVENT     & 1.00      & 1.00   & 1.00     & 3       \\
O         & 0.96      & 1.00   & 0.98     & 54      \\
GPE       & 1.00      & 1.00   & 1.00     & 2       \\
DATE      & 1.00      & 1.00   & 1.00     & 8       \\
FAC       & 1.00      & 1.00   & 1.00     & 8       \\
NORP      & 1.00      & 0.71   & 0.83     & 7       \\
LOC       & 1.00      & 1.00   & 1.00     & 13      \\
PERSON    & 1.00      & 1.00   & 1.00     & 8       \\
\midrule
accuracy  &           &        & 0.98     & 111     \\
macro avg & 1.00      & 0.97   & 0.98     & 111     \\
weighted avg & 0.98      & 0.98   & 0.98     & 111     \\
\bottomrule
\end{tabular}
\end{table}

\begin{table}[h]
\centering
\caption{English POS ALL for Evaluation}
\begin{tabular}{lrrrr}
\toprule
          & precision & recall & f1-score & support \\
\midrule
PART      & 1.00      & 1.00   & 1.00     & 7       \\
nan       & 1.00      & 1.00   & 1.00     & 19      \\
ADV       & 1.00      & 0.88   & 0.93     & 16      \\
SCONJ     & 0.88      & 1.00   & 0.93     & 7       \\
WDT       & 0.00      & 0.00   & 0.00     & 1       \\
PROPN     & 1.00      & 1.00   & 1.00     & 75      \\
PRON      & 1.00      & 1.00   & 1.00     & 7       \\
NUM       & 0.85      & 1.00   & 0.92     & 22      \\
PUNCT     & 1.00      & 1.00   & 1.00     & 48      \\
AUX       & 1.00      & 1.00   & 1.00     & 26      \\
ADP       & 1.00      & 1.00   & 1.00     & 62      \\
NOUN      & 1.00      & 1.00   & 1.00     & 89      \\
VERB      & 1.00      & 1.00   & 1.00     & 40      \\
CCONJ     & 0.93      & 1.00   & 0.96     & 26      \\
ADJ       & 1.00      & 0.93   & 0.97     & 30      \\
DET       & 1.00      & 0.95   & 0.98     & 44      \\
\midrule
accuracy  &           &        & 0.99     & 519     \\
macro avg & 0.92      & 0.92   & 0.92     & 519     \\
weighted avg & 0.99      & 0.99   & 0.99     & 519     \\
\bottomrule
\end{tabular}
\end{table}

\begin{table}[h]
\centering
\caption{English NER ALL for Evaluation}
\begin{tabular}{lrrrr}
\toprule
          & precision & recall & f1-score & support \\
\midrule
nan       & 1.00      & 1.00   & 1.00     & 19      \\
EVENT     & 1.00      & 1.00   & 1.00     & 7       \\
GPE       & 1.00      & 0.89   & 0.94     & 18      \\
O         & 0.97      & 1.00   & 0.98     & 283     \\
ORG       & 1.00      & 0.90   & 0.95     & 20      \\
MEASUREMENT & 1.00      & 1.00   & 1.00     & 2       \\
DATE      & 0.98      & 0.98   & 0.98     & 47      \\
FAC       & 1.00      & 1.00   & 1.00     & 17      \\
NORP      & 1.00      & 1.00   & 1.00     & 10      \\
MED       & 1.00      & 0.88   & 0.93     & 8       \\
FOOD      & 1.00      & 1.00   & 1.00     & 1       \\
LOC       & 1.00      & 0.93   & 0.97     & 45      \\
PERCENT   & 1.00      & 1.00   & 1.00     & 5       \\
WORK\_OF\_ART & 1.00      & 1.00   & 1.00     & 8       \\
PRODUCT   & 1.00      & 1.00   & 1.00     & 11      \\
PERSON    & 1.00      & 1.00   & 1.00     & 18      \\
\midrule
accuracy  &           &        & 0.98     & 519     \\
macro avg & 1.00      & 0.97   & 0.98     & 519     \\
weighted avg & 0.98      & 0.98   & 0.98     & 519     \\
\bottomrule
\end{tabular}
\end{table}

\begin{table}[h]
\centering
\caption{English POS PARALLEL for Evaluation}
\begin{tabular}{lrrrr}
\toprule
          & precision & recall & f1-score & support \\
\midrule
PART      & 1.00      & 1.00   & 1.00     & 3       \\
nan       & 1.00      & 1.00   & 1.00     & 16      \\
ADV       & 1.00      & 0.89   & 0.94     & 9       \\
SCONJ     & 0.75      & 1.00   & 0.86     & 3       \\
PROPN     & 1.00      & 1.00   & 1.00     & 35      \\
PRON      & 1.00      & 1.00   & 1.00     & 4       \\
NUM       & 0.85      & 1.00   & 0.92     & 11      \\
VERB      & 1.00      & 1.00   & 1.00     & 19      \\
PUNCT     & 1.00      & 1.00   & 1.00     & 25      \\
ADP       & 1.00      & 1.00   & 1.00     & 28      \\
NOUN      & 1.00      & 1.00   & 1.00     & 45      \\
CCONJ     & 1.00      & 1.00   & 1.00     & 12      \\
AUX       & 1.00      & 1.00   & 1.00     & 11      \\
ADJ       & 1.00      & 0.94   & 0.97     & 16      \\
DET       & 1.00      & 0.94   & 0.97     & 18      \\
\midrule
accuracy  &           &        & 0.99     & 255     \\
macro avg & 0.97      & 0.98   & 0.98     & 255     \\
weighted avg & 0.99      & 0.99   & 0.99     & 255     \\
\bottomrule
\end{tabular}
\end{table}

\begin{table}[h]
\centering
\caption{English NER PARALLEL for Evaluation}
\begin{tabular}{lrrrr}
\toprule
          & precision & recall & f1-score & support \\
\midrule
nan       & 1.00      & 1.00   & 1.00     & 16      \\
EVENT     & 1.00      & 1.00   & 1.00     & 3       \\
GPE       & 1.00      & 0.89   & 0.94     & 9       \\
O         & 0.96      & 0.99   & 0.97     & 134     \\
LOC       & 1.00      & 0.96   & 0.98     & 23      \\
MEASUREMENT & 1.00      & 1.00   & 1.00     & 1       \\
PERSON    & 1.00      & 1.00   & 1.00     & 9       \\
DATE      & 0.96      & 0.96   & 0.96     & 23      \\
FAC       & 1.00      & 1.00   & 1.00     & 7       \\
NORP      & 1.00      & 1.00   & 1.00     & 4       \\
MED       & 1.00      & 0.75   & 0.86     & 4       \\
FOOD      & 1.00      & 1.00   & 1.00     & 1       \\
ORG       & 1.00      & 0.80   & 0.89     & 10      \\
WORK\_OF\_ART & 1.00      & 1.00   & 1.00     & 4       \\
PRODUCT   & 1.00      & 1.00   & 1.00     & 5       \\
PERCENT   & 1.00      & 1.00   & 1.00     & 2       \\
\midrule
accuracy  &           &        & 0.97     & 255     \\
macro avg & 0.99      & 0.96   & 0.97     & 255     \\
weighted avg & 0.97      & 0.97   & 0.97     & 255     \\
\bottomrule
\end{tabular}
\end{table}

\begin{table}[h]
\centering
\caption{English POS PARALLEL HEALTH for Evaluation}
\begin{tabular}{lrrrr}
\toprule
          & precision & recall & f1-score & support \\
\midrule
PART      & 1.00      & 1.00   & 1.00     & 3       \\
nan       & 1.00      & 1.00   & 1.00     & 8       \\
ADV       & 1.00      & 0.83   & 0.91     & 6       \\
SCONJ     & 1.00      & 1.00   & 1.00     & 2       \\
PROPN     & 1.00      & 1.00   & 1.00     & 15      \\
NUM       & 0.80      & 1.00   & 0.89     & 8       \\
VERB      & 1.00      & 1.00   & 1.00     & 13      \\
PUNCT     & 1.00      & 1.00   & 1.00     & 15      \\
ADP       & 1.00      & 1.00   & 1.00     & 17      \\
NOUN      & 1.00      & 1.00   & 1.00     & 24      \\
CCONJ     & 1.00      & 1.00   & 1.00     & 6       \\
AUX       & 1.00      & 1.00   & 1.00     & 3       \\
ADJ       & 1.00      & 0.89   & 0.94     & 9       \\
DET       & 1.00      & 1.00   & 1.00     & 6       \\
\midrule
accuracy  &           &        & 0.99     & 135     \\
macro avg & 0.99      & 0.98   & 0.98     & 135     \\
weighted avg & 0.99      & 0.99   & 0.99     & 135     \\
\bottomrule
\end{tabular}
\end{table}

\begin{table}[h]
\centering
\caption{English NER PARALLEL HEALTH for Evaluation}
\begin{tabular}{lrrrr}
\toprule
          & precision & recall & f1-score & support \\
\midrule
nan       & 1.00      & 1.00   & 1.00     & 8       \\
GPE       & 1.00      & 0.83   & 0.91     & 6       \\
O         & 0.92      & 0.99   & 0.95     & 73      \\
LOC       & 1.00      & 0.86   & 0.92     & 7       \\
MEASUREMENT & 1.00      & 1.00   & 1.00     & 1       \\
PERSON    & 1.00      & 1.00   & 1.00     & 1       \\
DATE      & 0.92      & 0.92   & 0.92     & 13      \\
MED       & 1.00      & 0.75   & 0.86     & 4       \\
FOOD      & 1.00      & 1.00   & 1.00     & 1       \\
ORG       & 1.00      & 0.80   & 0.89     & 10      \\
WORK\_OF\_ART & 1.00      & 1.00   & 1.00     & 4       \\
PRODUCT   & 1.00      & 1.00   & 1.00     & 5       \\
PERCENT   & 1.00      & 1.00   & 1.00     & 2       \\
\midrule
accuracy  &           &        & 0.95     & 135     \\
macro avg & 0.99      & 0.93   & 0.96     & 135     \\
weighted avg & 0.95      & 0.95   & 0.95     & 135     \\
\bottomrule
\end{tabular}
\end{table}

\begin{table}[h]
\centering
\caption{English POS PARALLEL TOURISM for Evaluation}
\begin{tabular}{lrrrr}
\toprule
          & precision & recall & f1-score & support \\
\midrule
nan       & 1.00      & 1.00   & 1.00     & 6       \\
ADV       & 1.00      & 1.00   & 1.00     & 3       \\
SCONJ     & 0.50      & 1.00   & 0.67     & 1       \\
ADP       & 1.00      & 1.00   & 1.00     & 11      \\
PROPN     & 1.00      & 1.00   & 1.00     & 20      \\
CCONJ     & 1.00      & 1.00   & 1.00     & 6       \\
VERB      & 1.00      & 1.00   & 1.00     & 6       \\
AUX       & 1.00      & 1.00   & 1.00     & 8       \\
PUNCT     & 1.00      & 1.00   & 1.00     & 10      \\
NOUN      & 1.00      & 1.00   & 1.00     & 21      \\
PRON      & 1.00      & 1.00   & 1.00     & 4       \\
NUM       & 1.00      & 1.00   & 1.00     & 3       \\
ADJ       & 1.00      & 1.00   & 1.00     & 7       \\
DET       & 1.00      & 0.92   & 0.96     & 12      \\
\midrule
accuracy  &           &        & 0.99     & 118     \\
macro avg & 0.96      & 0.99   & 0.97     & 118     \\
weighted avg & 1.00      & 0.99   & 0.99     & 118     \\
\bottomrule
\end{tabular}
\end{table}

\begin{table}[h]
\centering
\caption{English NER PARALLEL TOURISM for Evaluation}
\begin{tabular}{lrrrr}
\toprule
          & precision & recall & f1-score & support \\
\midrule
nan       & 1.00      & 1.00   & 1.00     & 6       \\
EVENT     & 1.00      & 1.00   & 1.00     & 3       \\
O         & 1.00      & 1.00   & 1.00     & 61      \\
GPE       & 1.00      & 1.00   & 1.00     & 3       \\
DATE      & 1.00      & 1.00   & 1.00     & 10      \\
FAC       & 1.00      & 1.00   & 1.00     & 7       \\
NORP      & 1.00      & 1.00   & 1.00     & 4       \\
LOC       & 1.00      & 1.00   & 1.00     & 16      \\
PERSON    & 1.00      & 1.00   & 1.00     & 8       \\
\midrule
accuracy  &           &        & 1.00     & 118     \\
macro avg & 1.00      & 1.00   & 1.00     & 118     \\
weighted avg & 1.00      & 1.00   & 1.00     & 118     \\
\bottomrule
\end{tabular}
\end{table}

\begin{table}[h]
\centering
\caption{English POS PROMPT TOURISM for Evaluation}
\begin{tabular}{lrrrr}
\toprule
          & precision & recall & f1-score & support \\
\midrule
nan       & 1.00      & 1.00   & 1.00     & 7       \\
ADV       & 1.00      & 1.00   & 1.00     & 3       \\
SCONJ     & 0.50      & 1.00   & 0.67     & 1       \\
ADP       & 1.00      & 1.00   & 1.00     & 11      \\
PROPN     & 1.00      & 1.00   & 1.00     & 20      \\
CCONJ     & 1.00      & 1.00   & 1.00     & 6       \\
VERB      & 1.00      & 1.00   & 1.00     & 6       \\
AUX       & 1.00      & 1.00   & 1.00     & 8       \\
PUNCT     & 1.00      & 1.00   & 1.00     & 10      \\
NOUN      & 1.00      & 1.00   & 1.00     & 21      \\
PRON      & 1.00      & 1.00   & 1.00     & 4       \\
NUM       & 1.00      & 1.00   & 1.00     & 3       \\
ADJ       & 1.00      & 1.00   & 1.00     & 7       \\
DET       & 1.00      & 0.92   & 0.96     & 12      \\
\midrule
accuracy  &           &        & 0.99     & 119     \\
macro avg & 0.96      & 0.99   & 0.97     & 119     \\
weighted avg & 1.00      & 0.99   & 0.99     & 119     \\
\bottomrule
\end{tabular}
\end{table}

\begin{table}[h]
\centering
\caption{English NER PROMPT TOURISM for Evaluation}
\begin{tabular}{lrrrr}
\toprule
          & precision & recall & f1-score & support \\
\midrule
nan       & 1.00      & 1.00   & 1.00     & 7       \\
EVENT     & 1.00      & 1.00   & 1.00     & 3       \\
O         & 1.00      & 1.00   & 1.00     & 61      \\
GPE       & 1.00      & 1.00   & 1.00     & 3       \\
DATE      & 1.00      & 1.00   & 1.00     & 10      \\
FAC       & 1.00      & 1.00   & 1.00     & 7       \\
NORP      & 1.00      & 1.00   & 1.00     & 4       \\
LOC       & 1.00      & 1.00   & 1.00     & 16      \\
PERSON    & 1.00      & 1.00   & 1.00     & 8       \\
\midrule
accuracy  &           &        & 1.00     & 119     \\
macro avg & 1.00      & 1.00   & 1.00     & 119     \\
weighted avg & 1.00      & 1.00   & 1.00     & 119     \\
\bottomrule
\end{tabular}
\end{table}

\begin{table}[h]
\centering
\caption{English POS PROMPT for Evaluation}
\begin{tabular}{lrrrr}
\toprule
          & precision & recall & f1-score & support \\
\midrule
PART      & 1.00      & 1.00   & 1.00     & 4       \\
nan       & 1.00      & 1.00   & 1.00     & 16      \\
ADV       & 1.00      & 0.89   & 0.94     & 9       \\
SCONJ     & 0.80      & 1.00   & 0.89     & 4       \\
PROPN     & 1.00      & 1.00   & 1.00     & 35      \\
PRON      & 1.00      & 1.00   & 1.00     & 5       \\
NUM       & 0.85      & 1.00   & 0.92     & 11      \\
PUNCT     & 1.00      & 1.00   & 1.00     & 24      \\
AUX       & 1.00      & 1.00   & 1.00     & 13      \\
ADP       & 1.00      & 1.00   & 1.00     & 31      \\
NOUN      & 1.00      & 1.00   & 1.00     & 46      \\
VERB      & 1.00      & 1.00   & 1.00     & 21      \\
CCONJ     & 1.00      & 1.00   & 1.00     & 14      \\
ADJ       & 1.00      & 0.94   & 0.97     & 16      \\
DET       & 1.00      & 0.95   & 0.98     & 21      \\
\midrule
accuracy  &           &        & 0.99     & 270     \\
macro avg & 0.98      & 0.99   & 0.98     & 270     \\
weighted avg & 0.99      & 0.99   & 0.99     & 270     \\
\bottomrule
\end{tabular}
\end{table}

\begin{table}[h]
\centering
\caption{English NER PROMPT for Evaluation}
\begin{tabular}{lrrrr}
\toprule
          & precision & recall & f1-score & support \\
\midrule
nan       & 1.00      & 1.00   & 1.00     & 16      \\
EVENT     & 1.00      & 1.00   & 1.00     & 3       \\
GPE       & 1.00      & 0.89   & 0.94     & 9       \\
O         & 0.97      & 1.00   & 0.98     & 146     \\
ORG       & 1.00      & 0.80   & 0.89     & 10      \\
MEASUREMENT & 1.00      & 1.00   & 1.00     & 1       \\
DATE      & 1.00      & 1.00   & 1.00     & 24      \\
FAC       & 1.00      & 1.00   & 1.00     & 7       \\
NORP      & 1.00      & 1.00   & 1.00     & 4       \\
MED       & 1.00      & 1.00   & 1.00     & 4       \\
LOC       & 1.00      & 0.92   & 0.96     & 24      \\
PERCENT   & 1.00      & 1.00   & 1.00     & 3       \\
WORK\_OF\_ART & 1.00      & 1.00   & 1.00     & 4       \\
PRODUCT   & 1.00      & 1.00   & 1.00     & 6       \\
PERSON    & 1.00      & 1.00   & 1.00     & 9       \\
\midrule
accuracy  &           &        & 0.98     & 270     \\
macro avg & 1.00      & 0.97   & 0.98     & 270     \\
weighted avg & 0.98      & 0.98   & 0.98     & 270     \\
\bottomrule
\end{tabular}
\end{table}

\begin{table}[h]
\centering
\caption{English POS PROMPT HEALTH for Evaluation}
\begin{tabular}{lrrrr}
\toprule
          & precision & recall & f1-score & support \\
\midrule
PART      & 1.00      & 1.00   & 1.00     & 4       \\
nan       & 1.00      & 1.00   & 1.00     & 7       \\
ADV       & 1.00      & 0.83   & 0.91     & 6       \\
SCONJ     & 1.00      & 1.00   & 1.00     & 3       \\
PROPN     & 1.00      & 1.00   & 1.00     & 15      \\
PRON      & 1.00      & 1.00   & 1.00     & 1       \\
NUM       & 0.80      & 1.00   & 0.89     & 8       \\
PUNCT     & 1.00      & 1.00   & 1.00     & 14      \\
AUX       & 1.00      & 1.00   & 1.00     & 5       \\
ADP       & 1.00      & 1.00   & 1.00     & 20      \\
NOUN      & 1.00      & 1.00   & 1.00     & 25      \\
VERB      & 1.00      & 1.00   & 1.00     & 15      \\
CCONJ     & 1.00      & 1.00   & 1.00     & 8       \\
ADJ       & 1.00      & 0.89   & 0.94     & 9       \\
DET       & 1.00      & 1.00   & 1.00     & 9       \\
\midrule
accuracy  &           &        & 0.99     & 149     \\
macro avg & 0.99      & 0.98   & 0.98     & 149     \\
weighted avg & 0.99      & 0.99   & 0.99     & 149     \\
\bottomrule
\end{tabular}
\end{table}

\begin{table}[h]
\centering
\caption{English NER PROMPT HEALTH for Evaluation}
\begin{tabular}{lrrrr}
\toprule
          & precision & recall & f1-score & support \\
\midrule
nan       & 1.00      & 1.00   & 1.00     & 7       \\
GPE       & 1.00      & 0.83   & 0.91     & 6       \\
O         & 0.94      & 1.00   & 0.97     & 85      \\
ORG       & 1.00      & 0.80   & 0.89     & 10      \\
MEASUREMENT & 1.00      & 1.00   & 1.00     & 1       \\
DATE      & 1.00      & 1.00   & 1.00     & 14      \\
MED       & 1.00      & 1.00   & 1.00     & 4       \\
LOC       & 1.00      & 0.75   & 0.86     & 8       \\
PERCENT   & 1.00      & 1.00   & 1.00     & 3       \\
WORK\_OF\_ART & 1.00      & 1.00   & 1.00     & 4       \\
PRODUCT   & 1.00      & 1.00   & 1.00     & 6       \\
PERSON    & 1.00      & 1.00   & 1.00     & 1       \\
\midrule
accuracy  &           &        & 0.97     & 149     \\
macro avg & 1.00      & 0.95   & 0.97     & 149     \\
weighted avg & 0.97      & 0.97   & 0.96     & 149     \\
\bottomrule
\end{tabular}
\end{table}

\clearpage
\section{Detailed Evaluation Results (Figures)}
\label{sec:appendix_figures}

This appendix provides detailed evaluation results in the form of confusion matrices for both Bodo and English, covering POS tagging and NER, across various datasets (All that include both the Prompt and Parallel both in Health and Tourism, Parallel, Prompt, Health and Tourism).  Each image shows the True values and predicted values for each class. The images are presented side-by-side for easy comparison where appropriate (Bodo vs. English, POS vs. NER for the same dataset configuration).
\begin{figure}[htbp]
    \centering
    
    \begin{subfigure}{0.48\textwidth}
        \centering
        \includegraphics[width=\linewidth]{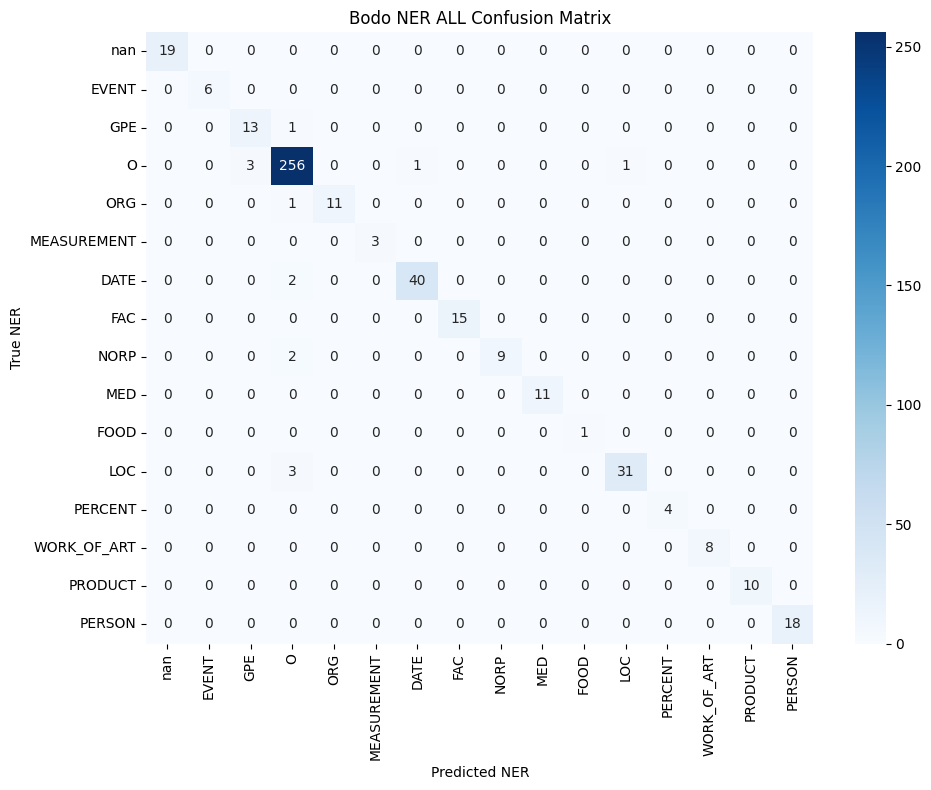}  
        \caption{Bodo NER (All Data)}
        \label{fig:bodo_ner_all}
    \end{subfigure}
    \hfill
    \begin{subfigure}{0.48\textwidth}
        \centering
        \includegraphics[width=\linewidth]{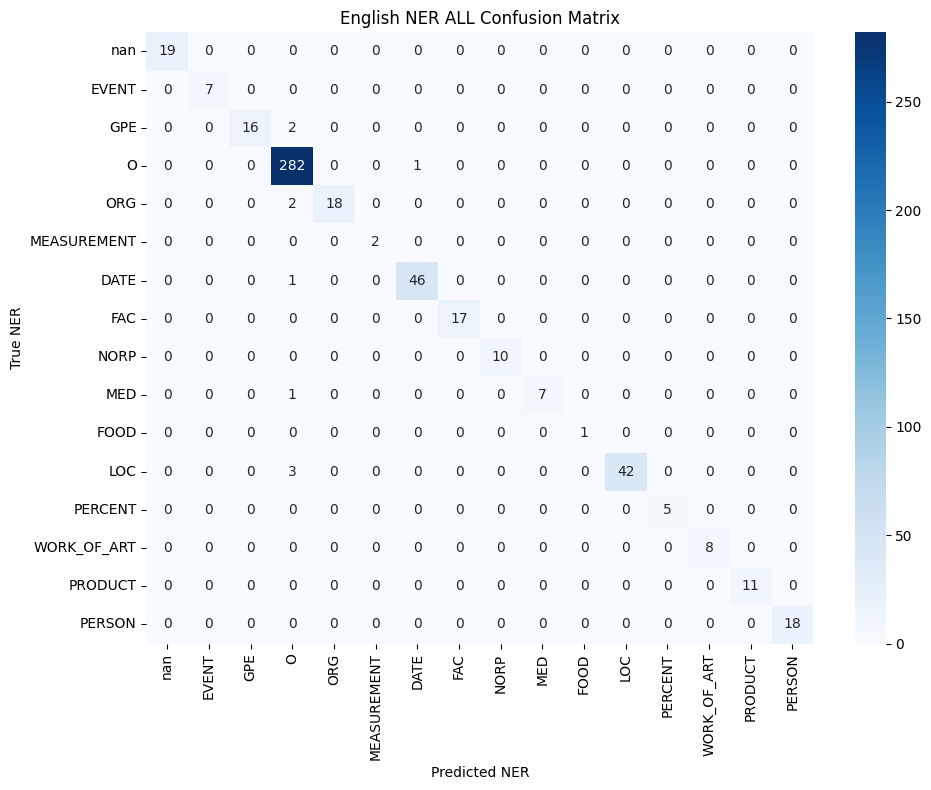}
        \caption{English NER (All Data)}
        \label{fig:english_ner_all}
    \end{subfigure}

    \begin{subfigure}{0.48\textwidth}
        \centering
        \includegraphics[width=\linewidth]{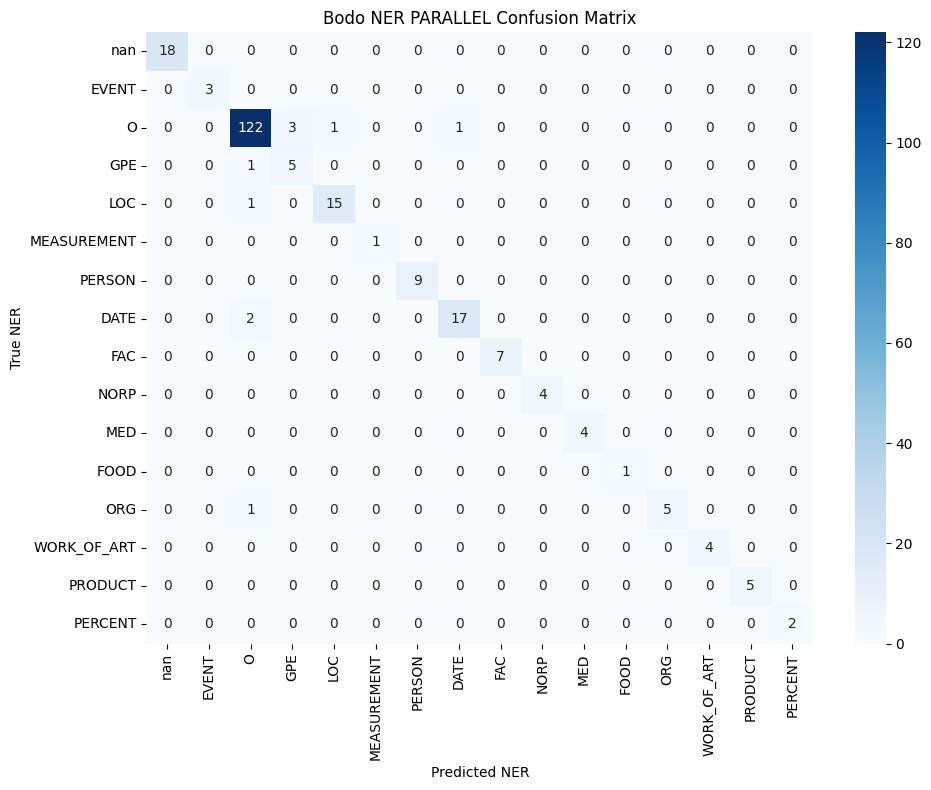}
        \caption{Bodo NER (Parallel Data)}
        \label{fig:bodo_ner_parallel}
    \end{subfigure}
    \hfill
    \begin{subfigure}{0.48\textwidth}
        \centering
        \includegraphics[width=\linewidth]{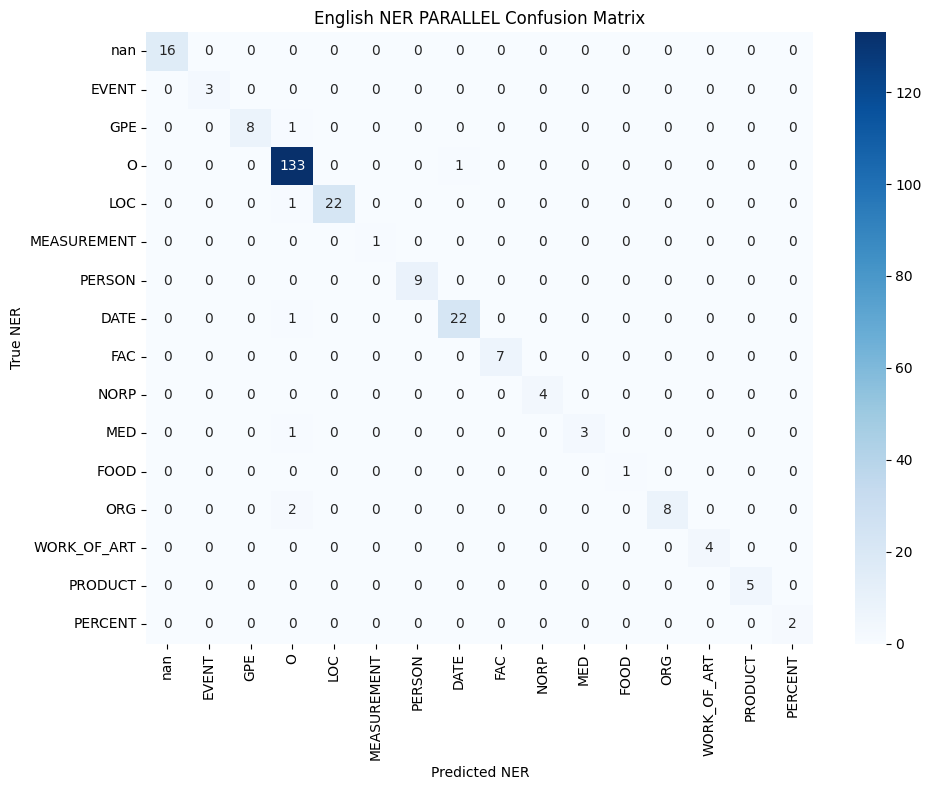}
        \caption{English NER (Parallel Data)}
        \label{fig:english_ner_parallel}
    \end{subfigure}
    
    \begin{subfigure}{0.48\textwidth}
        \centering
        \includegraphics[width=\linewidth]{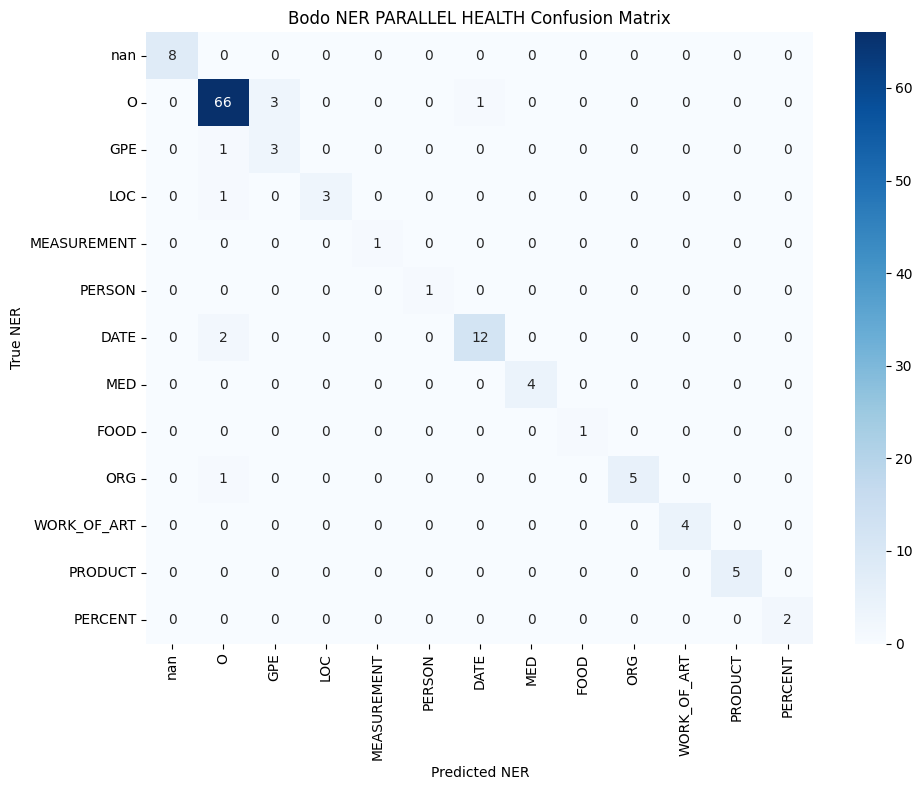}
        \caption{Bodo NER (Parallel Health)}
        \label{fig:bodo_ner_parallel_health}
    \end{subfigure}
    \hfill
    \begin{subfigure}{0.48\textwidth}
        \centering
        \includegraphics[width=\linewidth]{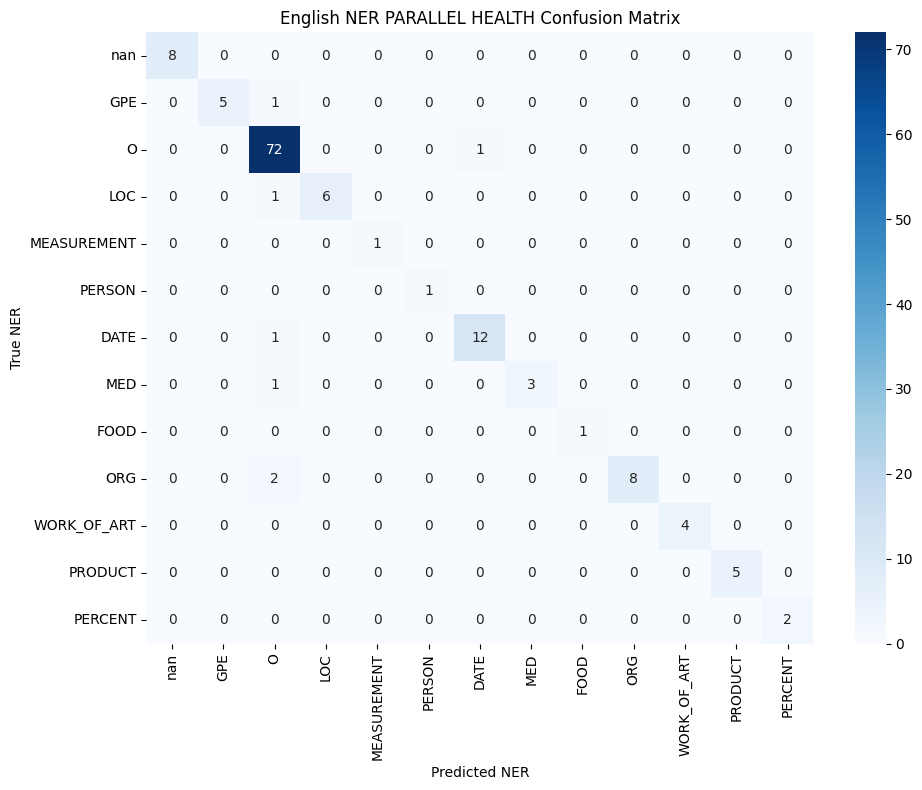}
        \caption{English NER (Parallel Health)}
        \label{fig:english_ner_parallel_health}
    \end{subfigure}

    \begin{subfigure}{0.48\textwidth}
        \centering
        \includegraphics[width=\linewidth]{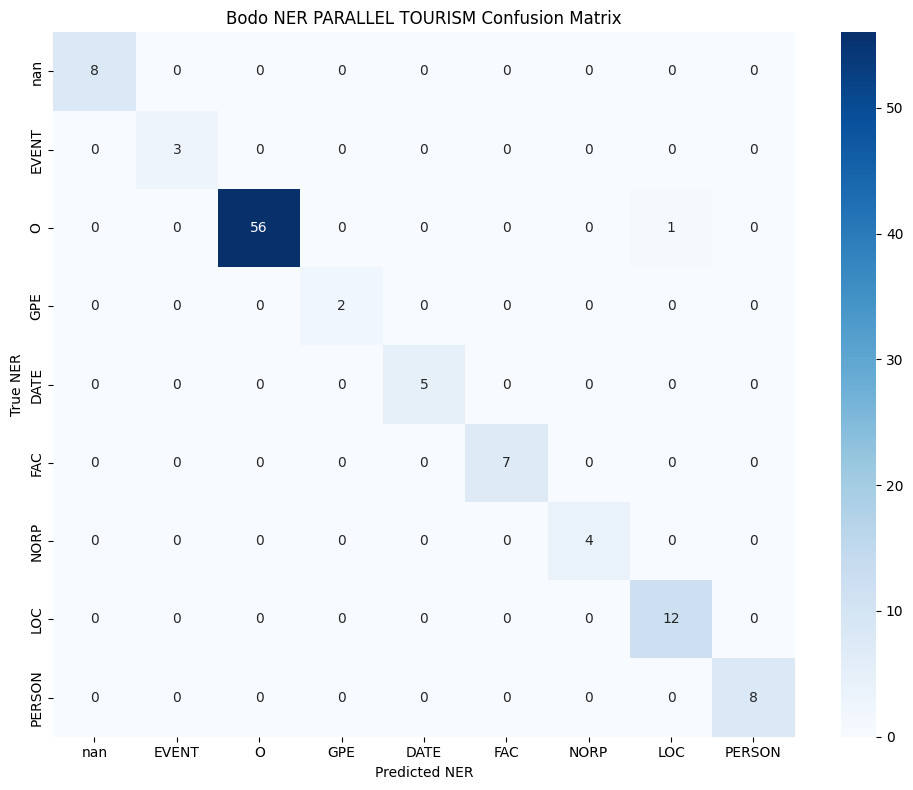}
        \caption{Bodo NER (Parallel Tourism)}
        \label{fig:bodo_ner_parallel_tourism}
    \end{subfigure}
    \hfill
    \begin{subfigure}{0.48\textwidth}
        \centering
        \includegraphics[width=\linewidth]{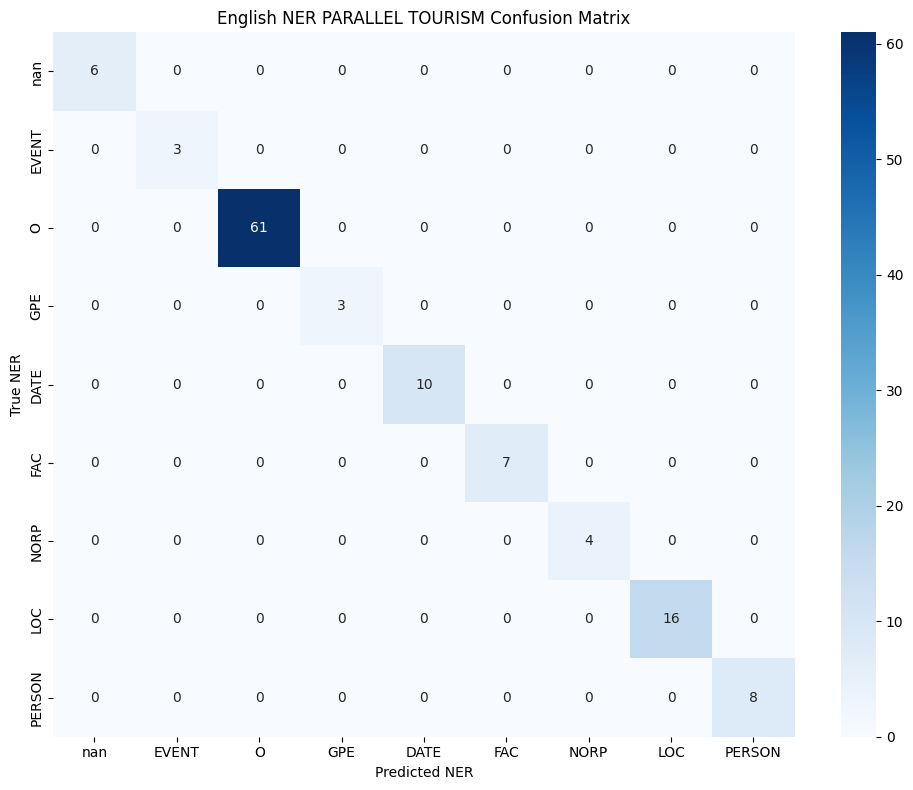}
        \caption{English NER (Parallel Tourism)}
        \label{fig:english_ner_parallel_tourism}
    \end{subfigure}

    \caption{Comparison of NER results for Bodo and English (Part 1).}
    \label{fig:ner_results_part1}
\end{figure}

\afterpage{%
\begin{figure}[htbp]
    \centering
    \begin{subfigure}{0.48\textwidth}
         \centering
        \includegraphics[width=\linewidth]{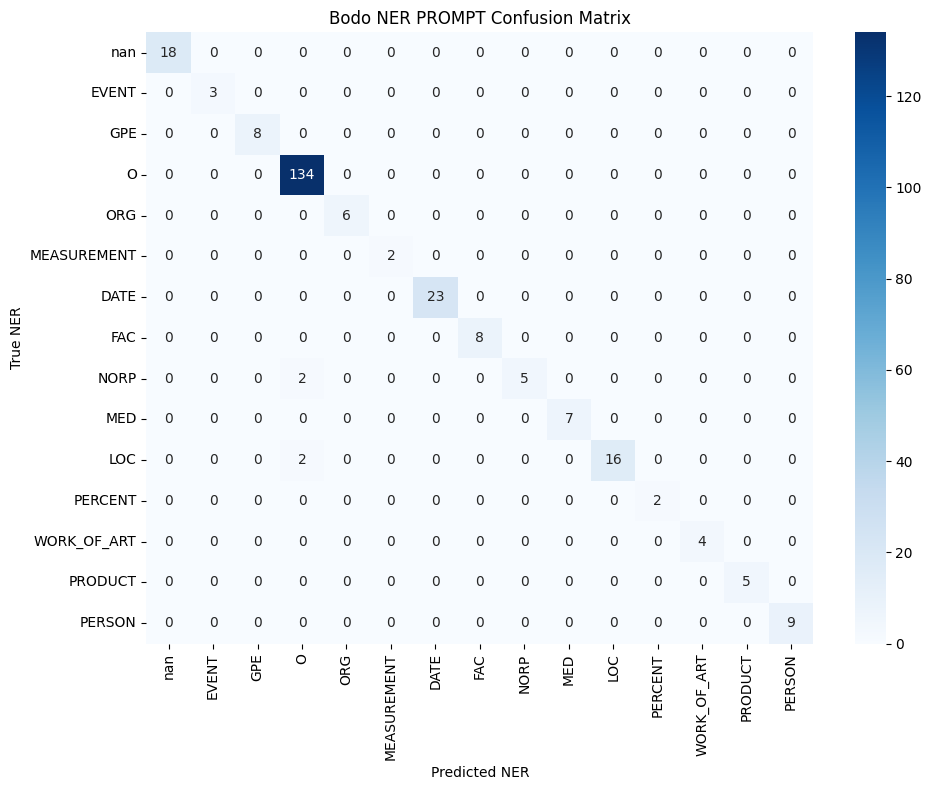}
        \caption{Bodo NER (Prompt-based)}
        \label{fig:bodo_ner_prompt}
    \end{subfigure}
    \hfill
    \begin{subfigure}{0.48\textwidth}
        \centering
        \includegraphics[width=\linewidth]{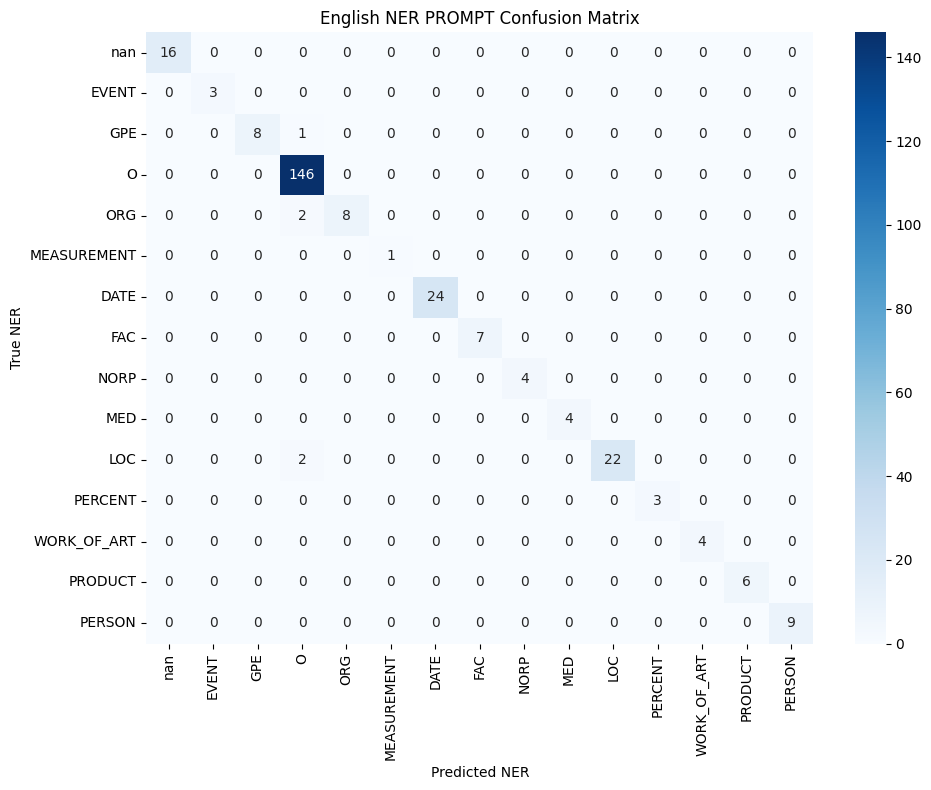}
        \caption{English NER (Prompt-based)}
        \label{fig:english_ner_prompt}
    \end{subfigure}

    \begin{subfigure}{0.48\textwidth}
        \centering
        \includegraphics[width=\linewidth]{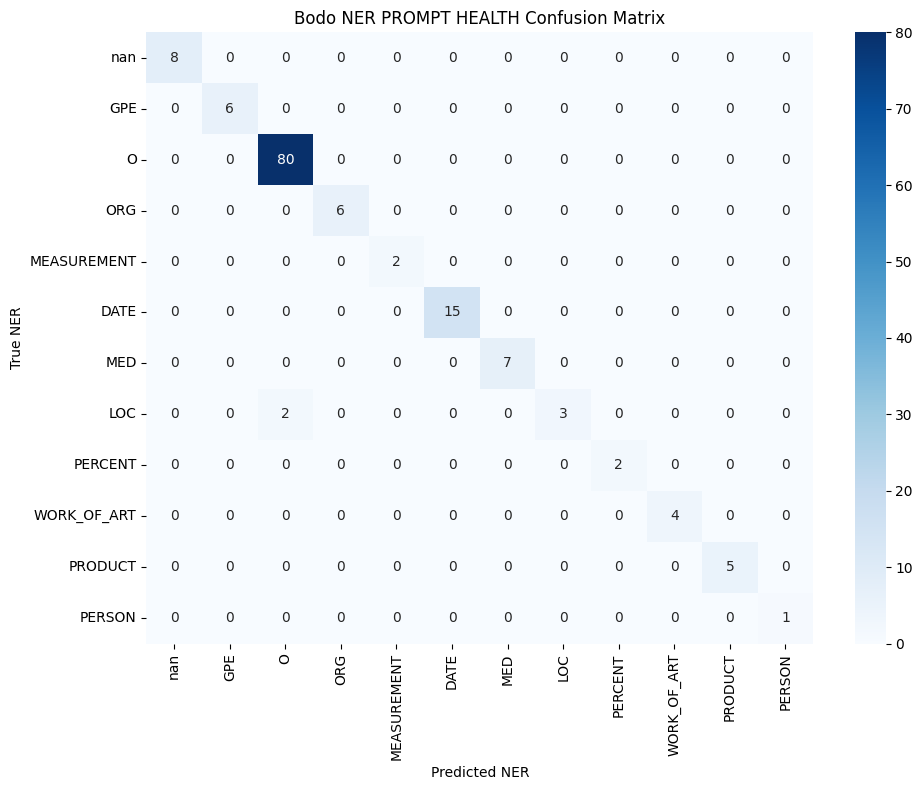}
        \caption{Bodo NER (Prompt Health)}
        \label{fig:bodo_ner_prompt_health}
    \end{subfigure}
    \hfill
    \begin{subfigure}{0.48\textwidth}
        \centering
        \includegraphics[width=\linewidth]{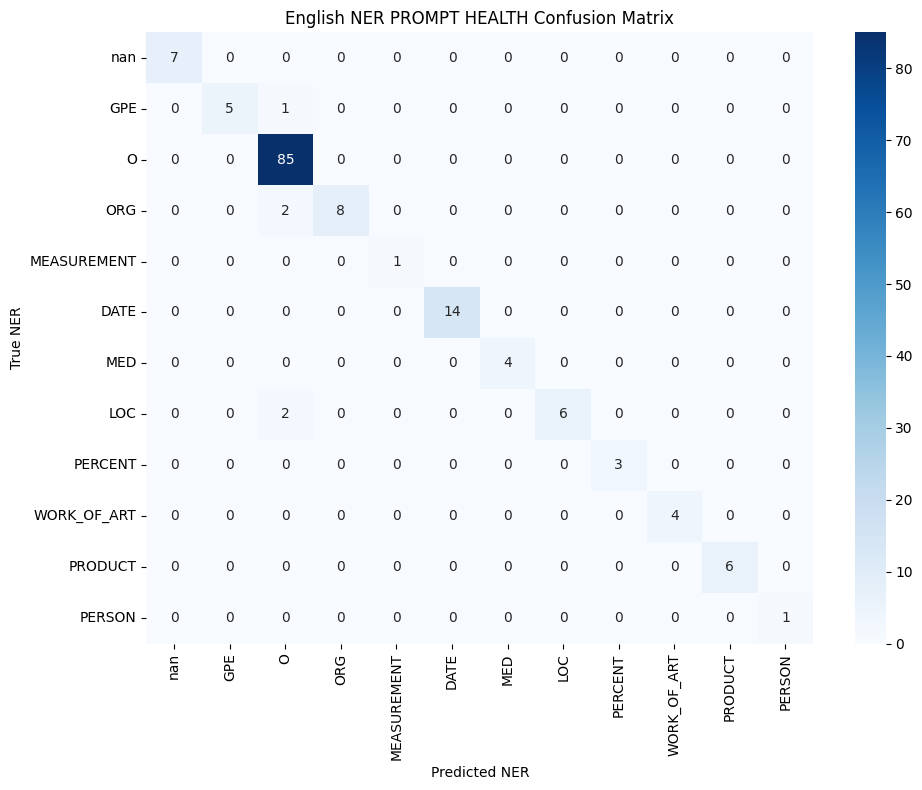}
        \caption{English NER (Prompt Health)}
        \label{fig:english_ner_prompt_health}
    \end{subfigure}

     \begin{subfigure}{0.48\textwidth}
        \centering
        \includegraphics[width=\linewidth]{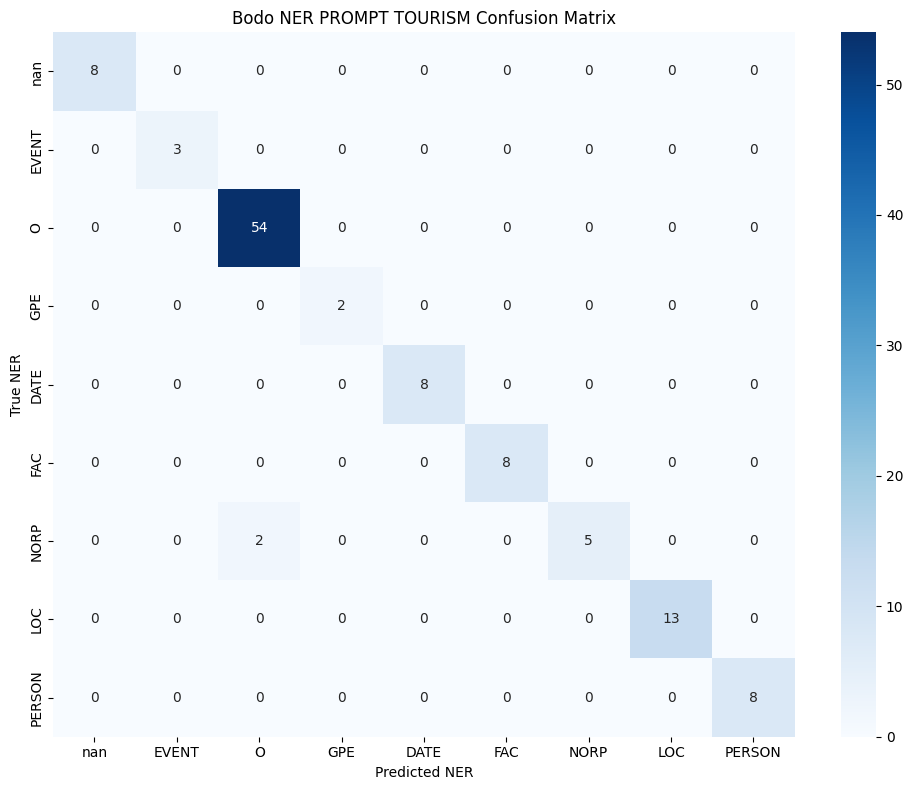}
        \caption{Bodo NER (Prompt Tourism)}
        \label{fig:bodo_ner_prompt_tourism}
    \end{subfigure}
    \hfill
    \begin{subfigure}{0.48\textwidth}
        \centering
        \includegraphics[width=\linewidth]{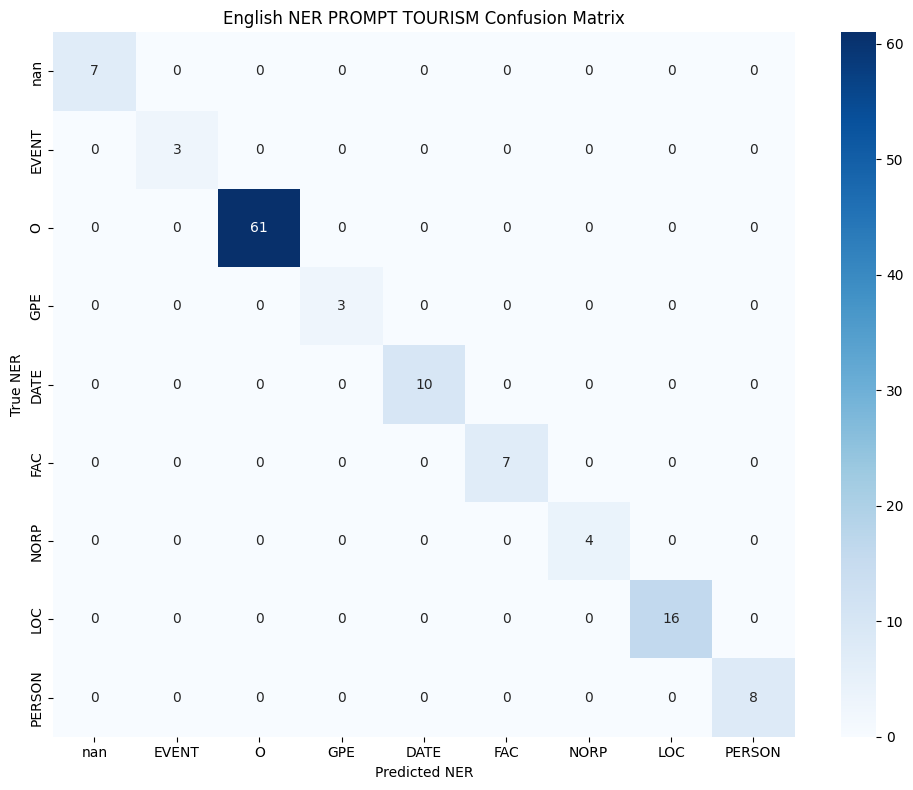}
        \caption{English NER (Prompt Tourism)}
        \label{fig:english_ner_prompt_tourism}
    \end{subfigure}
     \caption{Comparison of NER results for Bodo and English (Part 2).}
    \label{fig:ner_results_part2}
\end{figure}
} 

\afterpage{
\begin{figure}[htbp]
    \centering

    \begin{subfigure}{0.48\textwidth}
        \centering
        \includegraphics[width=\linewidth]{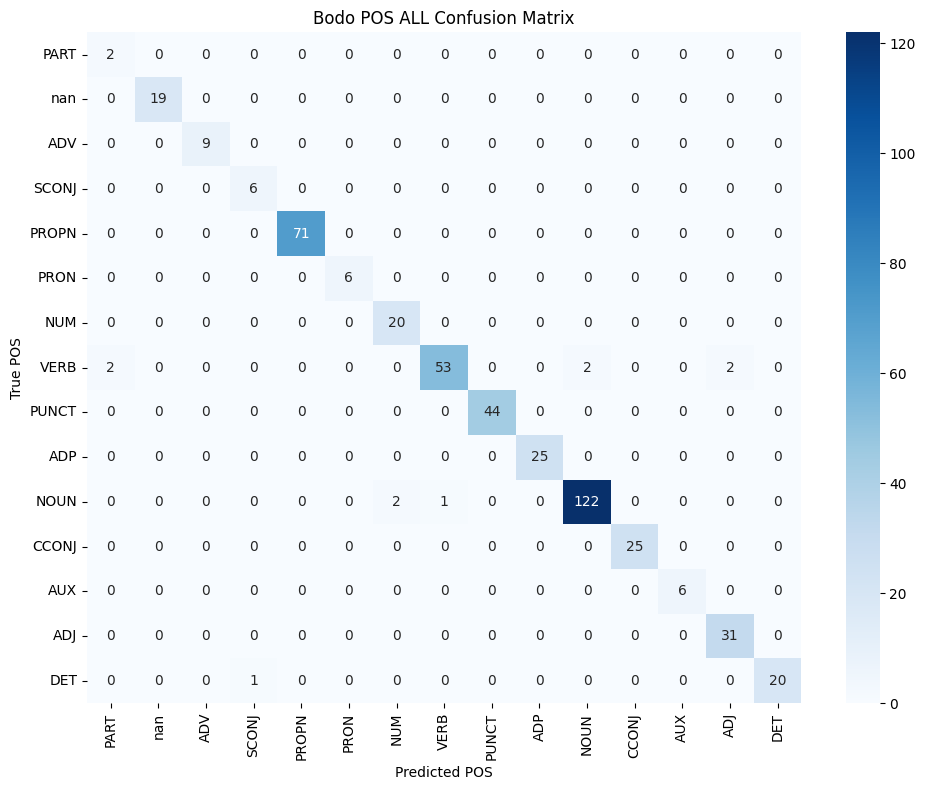}
        \caption{Bodo POS (All Data)}
        \label{fig:bodo_pos_all}
    \end{subfigure}
    \hfill
    \begin{subfigure}{0.48\textwidth}
        \centering
        \includegraphics[width=\linewidth]{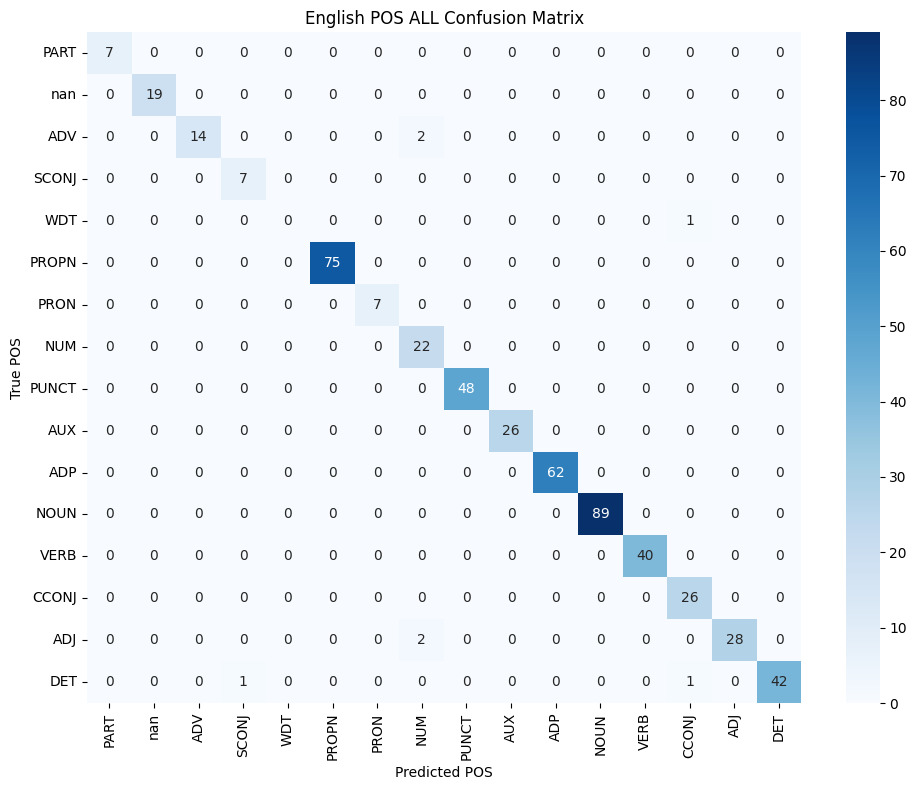}
        \caption{English POS (All Data)}
        \label{fig:english_pos_all}
    \end{subfigure}

    \begin{subfigure}{0.48\textwidth}
       \centering
        \includegraphics[width=\linewidth]{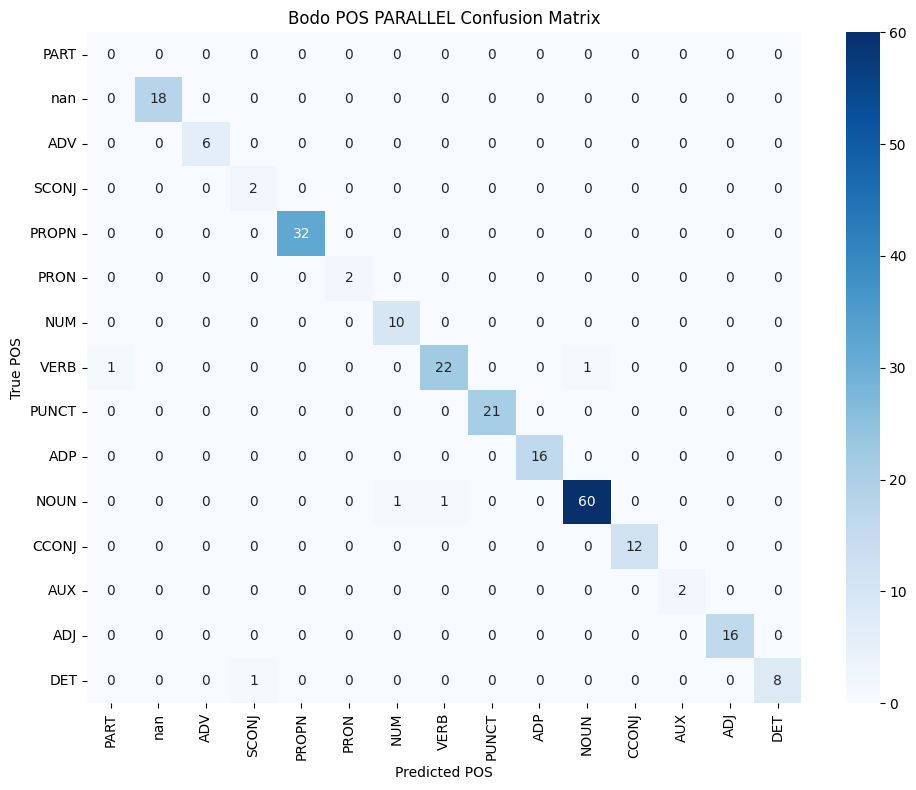}
        \caption{Bodo POS (Parallel Data)}
        \label{fig:bodo_pos_parallel}
    \end{subfigure}
    \hfill
    \begin{subfigure}{0.48\textwidth}
        \centering
        \includegraphics[width=\linewidth]{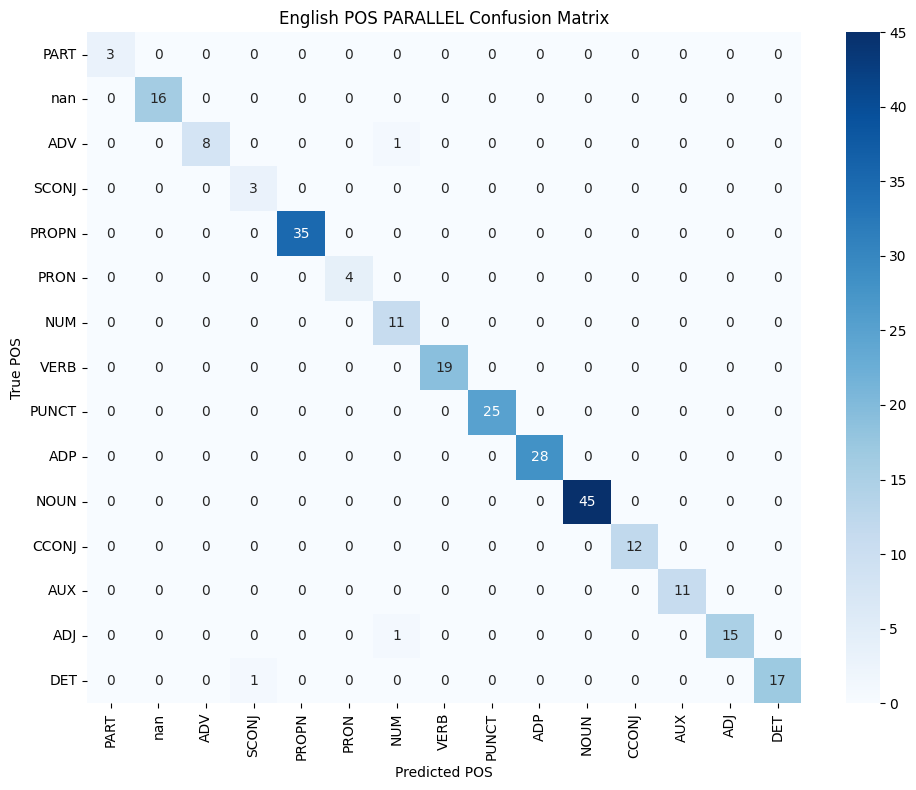}
        \caption{English POS (Parallel Data)}
        \label{fig:english_pos_parallel}
    \end{subfigure}

     \begin{subfigure}{0.48\textwidth}
        \centering
        \includegraphics[width=\linewidth]{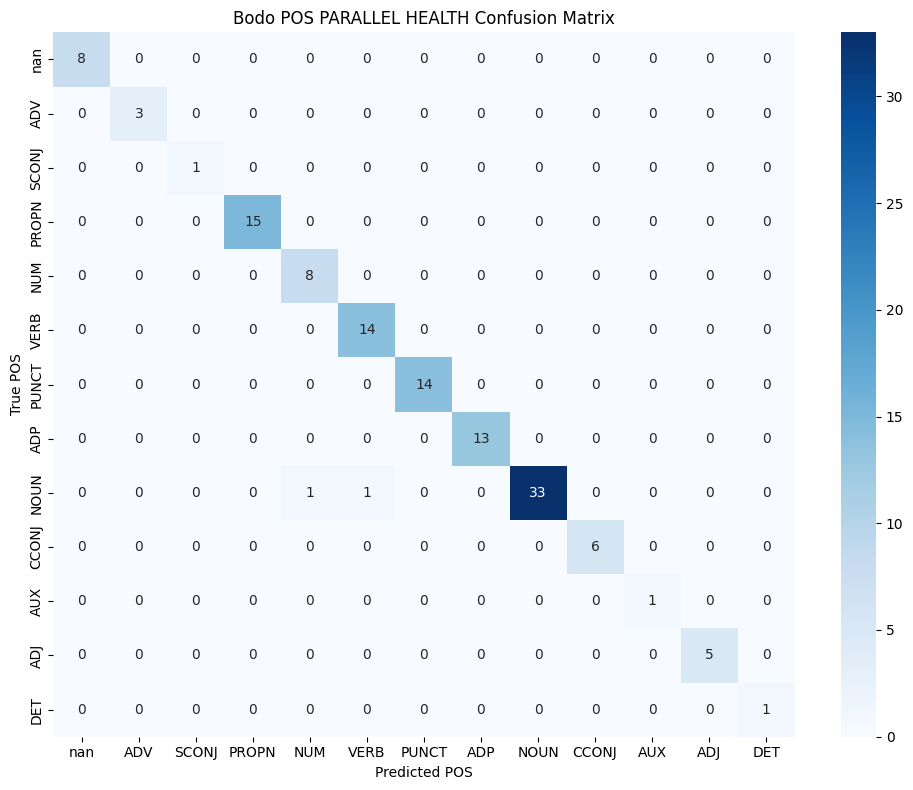}
        \caption{Bodo POS (Parallel Health)}
        \label{fig:bodo_pos_parallel_health}
    \end{subfigure}
    \hfill
    \begin{subfigure}{0.48\textwidth}
        \centering
        \includegraphics[width=\linewidth]{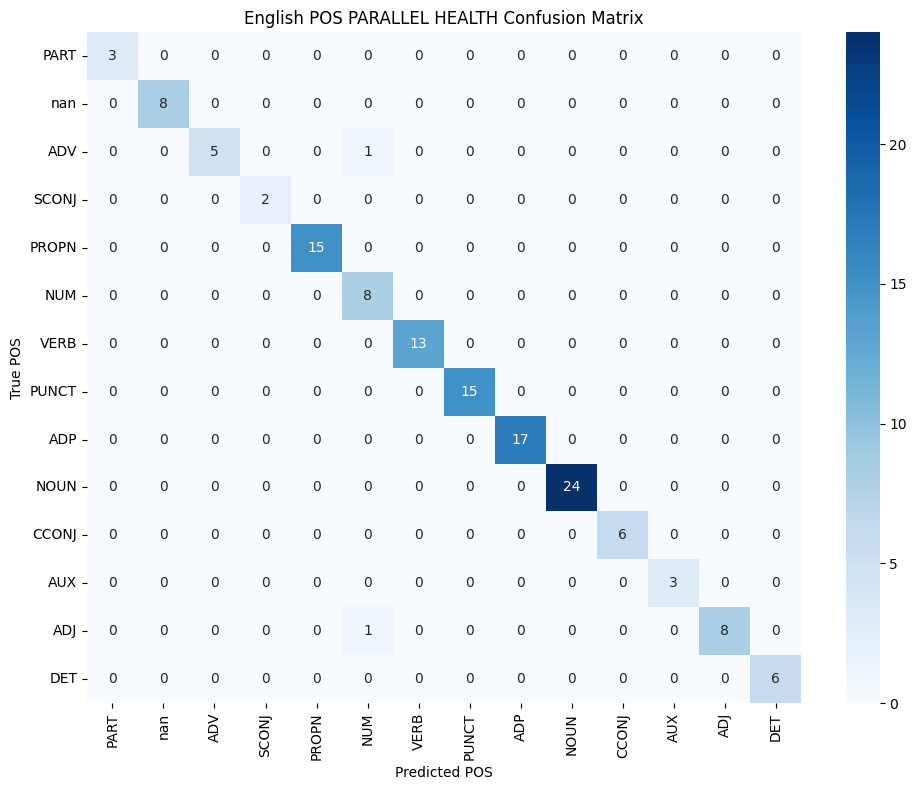}
        \caption{English POS (Parallel Health)}
        \label{fig:english_pos_parallel_health}
    \end{subfigure}
     \caption{Comparison of POS tagging results for Bodo and English (Part 1).}
    \label{fig:pos_results_part1}

\end{figure}
} 

\afterpage{
\begin{figure}[htbp]
      \begin{subfigure}{0.48\textwidth}
          \centering
          \includegraphics[width=\linewidth]{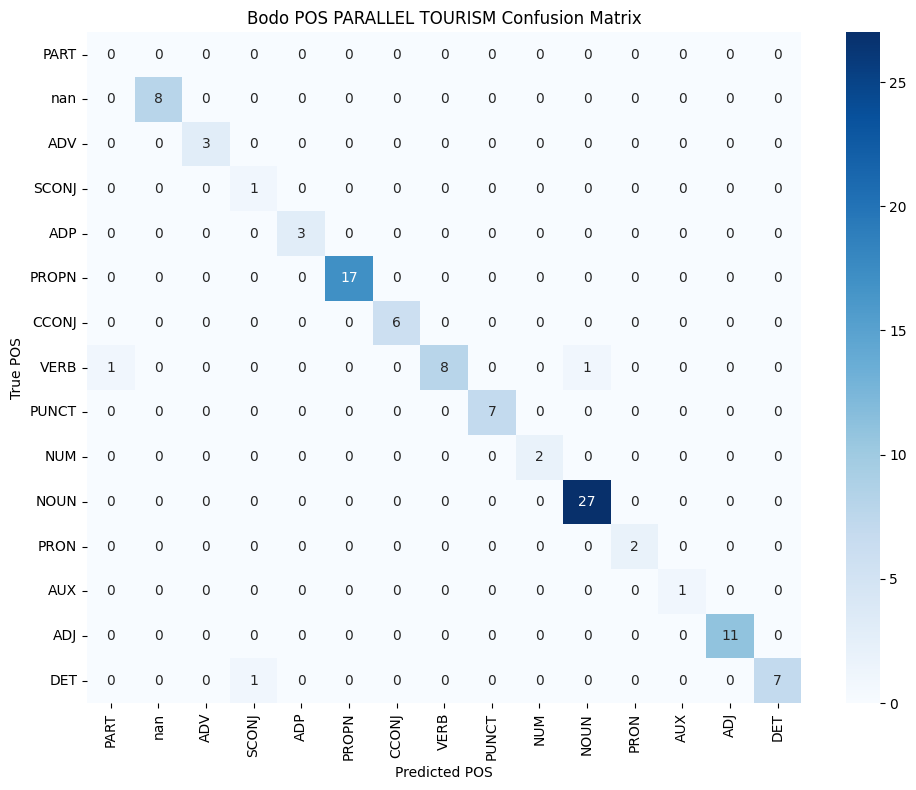}
          \caption{Bodo POS (Parallel Tourism)}
          \label{fig:bodo_pos_parallel_tourism}
      \end{subfigure}
      \hfill
      \begin{subfigure}{0.48\textwidth}
          \centering
          \includegraphics[width=\linewidth]{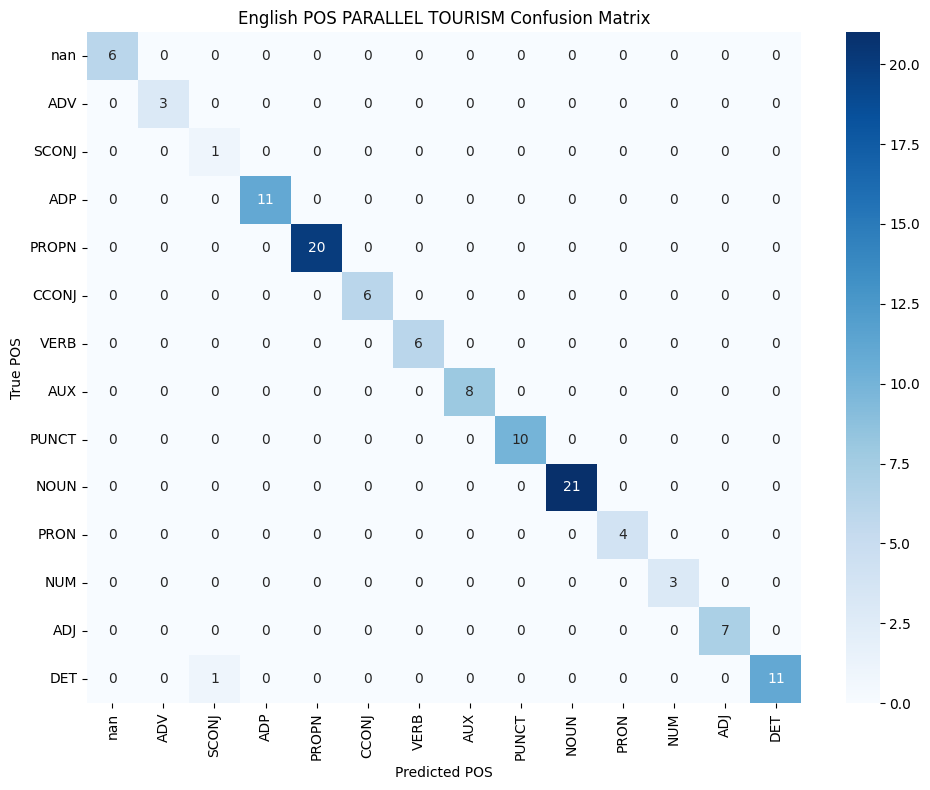}
          \caption{English POS (Parallel Tourism)}
          \label{fig:english_pos_parallel_tourism}
      \end{subfigure}
    
     \begin{subfigure}{0.48\textwidth}
          \centering
          \includegraphics[width=\linewidth]{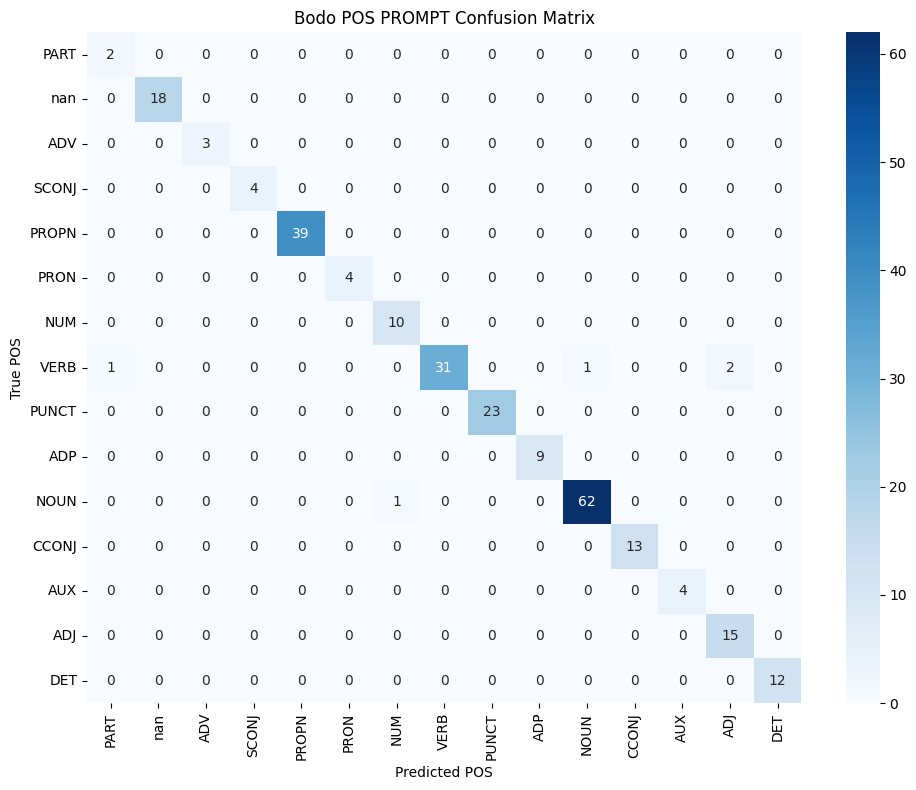}
          \caption{Bodo POS (Prompt-based)}
          \label{fig:bodo_pos_prompt}
      \end{subfigure}
      \hfill
      \begin{subfigure}{0.48\textwidth}
          \centering
          \includegraphics[width=\linewidth]{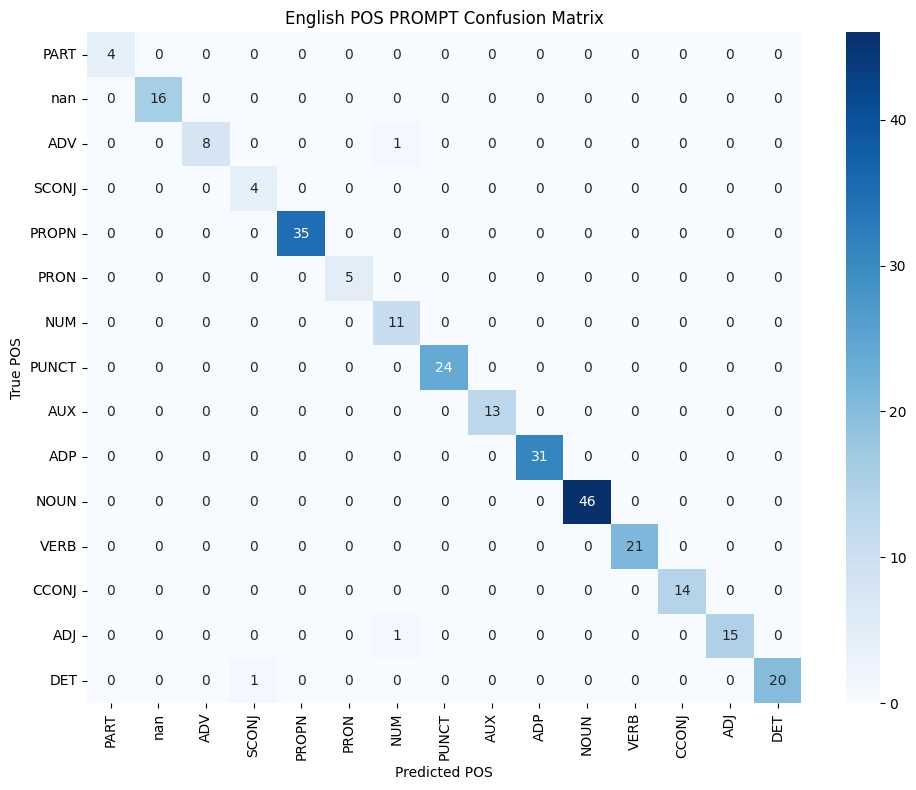}
          \caption{English POS (Prompt-based)}
          \label{fig:english_pos_prompt}
      \end{subfigure}

     \begin{subfigure}{0.48\textwidth}
          \centering
          \includegraphics[width=\linewidth]{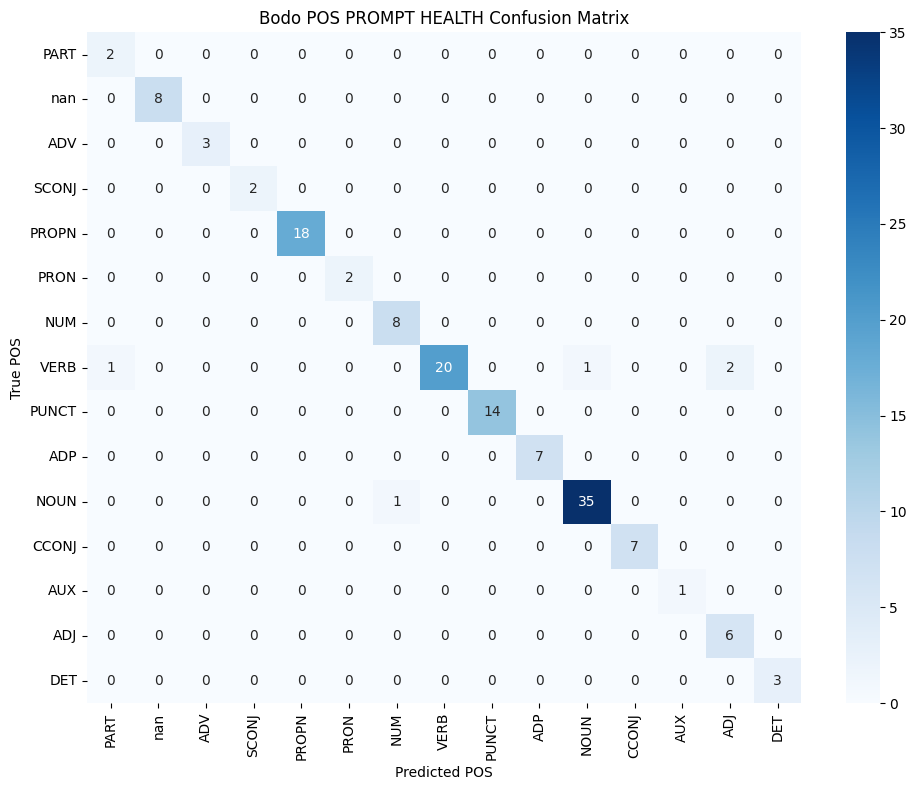}
          \caption{Bodo POS (Prompt Health)}
          \label{fig:bodo_pos_prompt_health}
      \end{subfigure}
      \hfill
      \begin{subfigure}{0.48\textwidth}
          \centering
          \includegraphics[width=\linewidth]{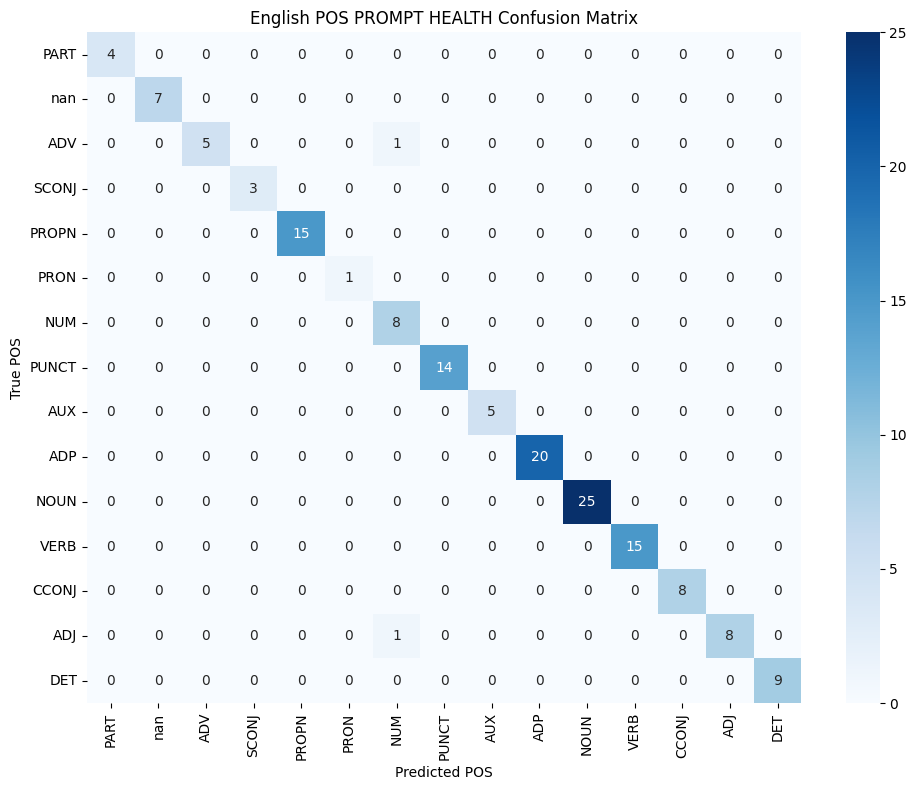}
          \caption{English POS (Prompt Health)}
          \label{fig:english_pos_prompt_health}
      \end{subfigure}
    
     \begin{subfigure}{0.48\textwidth}
        \centering
        \includegraphics[width=\linewidth]{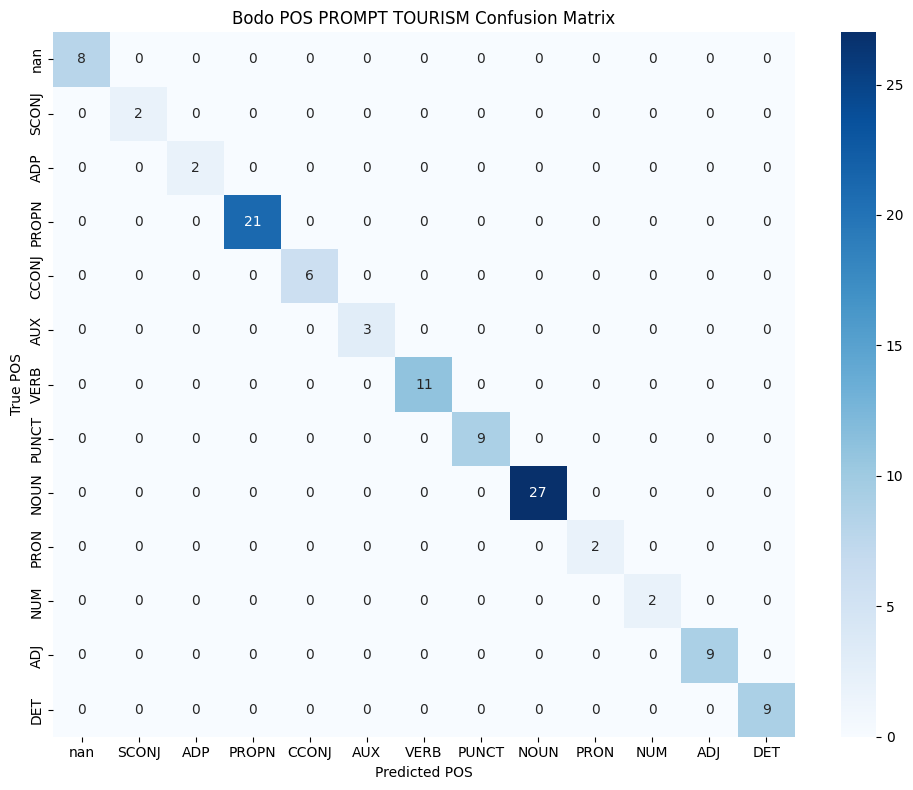}
        \caption{Bodo POS (Prompt Tourism)}
        \label{fig:bodo_pos_prompt_tourism}
    \end{subfigure}
    \hfill
    \begin{subfigure}{0.48\textwidth}
        \centering
        \includegraphics[width=\linewidth]{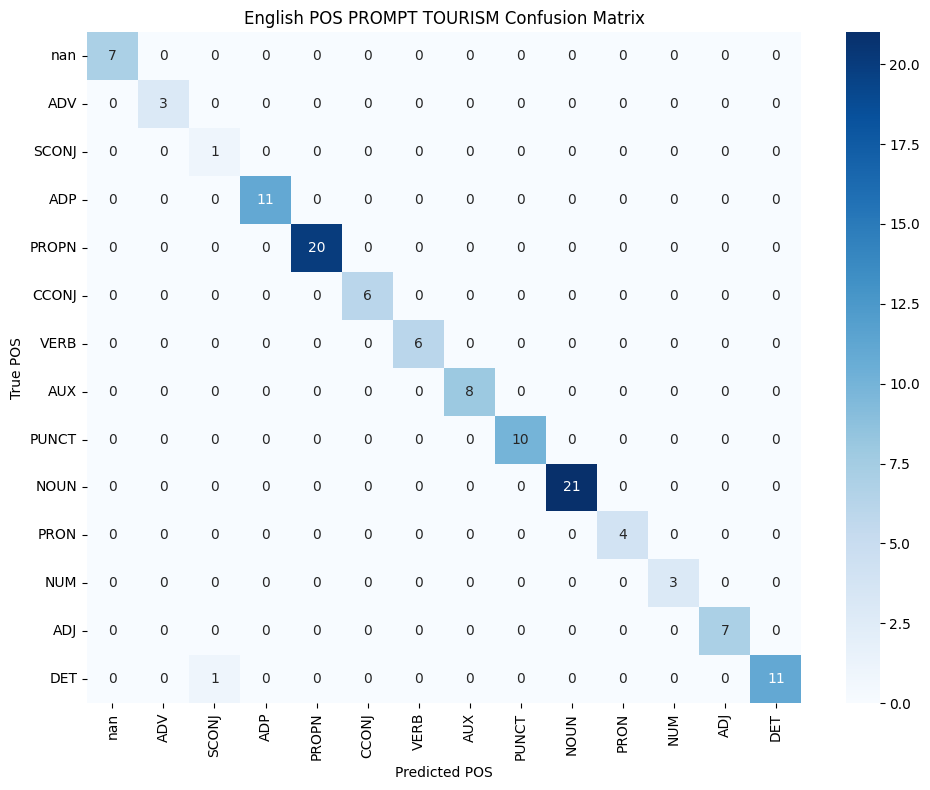}
        \caption{English POS (Prompt Tourism)}
        \label{fig:english_pos_prompt_tourism}
    \end{subfigure}

    \caption{Comparison of POS tagging results for Bodo and English (Part 2).}
    \label{fig:pos_results_part2}
\end{figure}
} 



\end{appendices}


\end{document}